\documentclass{article} % For LaTeX2e
\usepackage{iclr2023_conference,times}

\newcommand{\Appendix}[1]{the full version for}

\newcommand{\x}{\bm{x}}

\newcommand{\z}{\bm{z}}

\newcommand{\I}{\bm{I}}

\newcommand{\R}{\mathbb{R}}
\renewcommand{\Re}{\mathbb{R}}

\newcommand{\X}{\bm{X}}

\newcommand{\Z}{\bm{Z}}

\newcommand{\tr}{\textsf{tr}}

\renewcommand{\mathbf}{\boldsymbol}

\usepackage{hyperref}
\usepackage{url}
\usepackage[utf8]{inputenc} % allow utf-8 input
\usepackage[T1]{fontenc}    % use 8-bit T1 fonts
\usepackage{hyperref}       % hyperlinks
\usepackage{url}            % simple URL typesetting
\usepackage{booktabs}       % professional-quality tables
\usepackage{amsfonts}       % blackboard math symbols
\usepackage{nicefrac}       % compact symbols for 1/2, etc.
\usepackage{microtype}      % microtypography
\usepackage{bbding}
\usepackage{pifont}

\usepackage{color}

\usepackage{bm}
\usepackage[normalem]{ulem}
\usepackage{times,epsfig,graphics,graphicx,subfigure,wrapfig,float,caption,booktabs,xcolor,multirow,multicol,array,color,ifthen,tabu,colortbl,url,xparse,mathtools,patchcmd,enumitem,amssymb,xspace,nicefrac,microtype,amsmath,amsfonts,amsthm}
\newcommand{\compareyes}{{\color{green} \ding{52}}}
\newcommand{\compareno}{{\color{red} \ding{55}}}
\newcommand{\ours}{U-CTRL}

\title{Unsupervised Learning of Structured Representations via Closed-Loop Transcription}

\author{
\centerline{
Shengbang Tong\textsuperscript{\rm 1}\thanks{Equal contribution} \quad
Xili Dai\textsuperscript{\rm 1,2 *}\thanks{Work done during visiting at Berkeley} \quad
Yubei Chen\textsuperscript{\rm 3} \quad
Mingyang Li\textsuperscript{\rm 5} \quad
Zengyi Li\textsuperscript{\rm 1} \quad
Brent Yi\textsuperscript{\rm 1} \quad
} \\
\centerline{
\textbf{Yann LeCun}\textsuperscript{\rm 3,4}\quad
\textbf{Yi Ma}\textsuperscript{\rm 1,5}
}\\
\centerline{
\textsuperscript{\rm 1}University of California, Berkeley  \quad
\textsuperscript{\rm 2}Hong Kong University of Science and Technology (Guangzhou)}\\
\centerline{
\textsuperscript{\rm 3}Center for Data Science, New York University\quad
\textsuperscript{\rm 4}Courant Inst., New York University
}\\
\centerline{
\textsuperscript{\rm 5}Tsinghua-Berkeley Shenzhen Institute (TBSI) \quad 
}
}

\iclrfinalcopy
\begin{document}

\maketitle

\vspace{-4mm}   
\begin{abstract}
\vspace{-2mm}
This paper proposes an unsupervised method for learning a unified representation that serves both discriminative and generative purposes. While most existing unsupervised learning approaches focus on a representation for only one of these two goals, we show that a unified representation can enjoy the mutual benefits of having both. Such a representation is attainable by generalizing the recently proposed \textit{closed-loop transcription} framework, known as CTRL, to the unsupervised setting. This entails solving a constrained maximin game over a rate reduction objective that expands features of all samples while compressing features of augmentations of each sample. Through this process, we see discriminative low-dimensional structures emerge in the resulting representations.  Under comparable experimental conditions and network complexities, we demonstrate that these structured representations enable classification performance close to state-of-the-art unsupervised discriminative representations, and conditionally generated image quality significantly higher than that of state-of-the-art unsupervised generative models. Source code can be found at \href{https://github.com/Delay-Xili/uCTRL}{\textsc{}{https://github.com/Delay-Xili/uCTRL}}.

\end{abstract}

\vspace{-4mm}
\section{Introduction}
\vspace{-3mm}
In the past decade, we have witnessed an explosive development in the practice of machine learning, particularly with deep learning methods. 
% BY: given the state of NLP etc, seems hard to argue that successful applications "largely rely on hand-labeled data"
% Nonetheless, successful applications largely rely on training deep neural networks supervised with hand-labeled data, which can be costly and time-consuming to prepare.
% This has hindered progress in many application domains.
A key driver of success in practical applications 
%has been the combination of large datasets and scale; this has hindered progress in application domains that rely on supervision with costly hand-labeled data.
has been marvelous engineering endeavors, often focused on fitting increasingly large deep networks to input data paired with task-specific sets of labels.
Brute-force approaches of this nature, however, exert tremendous demands on hand-labeled data for supervision and computational resources for training and inference.
As a result, an increasing amount of attention has been directed toward using self-supervised or unsupervised techniques to learn representations that can not only learn without human annotation effort, but also be shared across downstream tasks. %from unlabelled data, say via self-supervised or unsupervised learning, to facilitate downstream tasks,  has drawn increasing attention.

%In last decade, we witness an explosive development in deep learning field. Industry and academy were enjoying the benefits, but also paying the "price", the procedure of labelling the huge number data which is time/money-consuming. Learning a deep learning model without human label becomes urgently for the whole field, especially for industry. Hence, unsupervised learning becomes more and more important.

\textbf{Discriminative versus Generative.}
Tasks in unsupervised learning are typically separated into two categories.
% Discriminative ones aim to extract low-dimensional class or semantic information from high-dimensional observations, and generative ones aim to enable sampling, often with semantically meaningful conditioning, from high-dimensional observation spaces.
Discriminative ones frame high-dimensional observations as inputs, from which low-dimensional class or latent information can be extracted, while generative ones frame observations as generated outputs, which should often be sampled given some semantically meaningful conditioning.

% An abundance of unsupervised learning methods have been proposed to produce representations amenable to discriminative downstream tasks.
Unsupervised learning approaches targeted at discriminative tasks are mainly based on a key idea: to pull different views from the same instance closer while enforcing a non-collapsed representation by
either contrastive learning techniques \citep{chen2020simple,he2020momentum,grill2020bootstrap}, covariance regularization methods \citep{bardes2021vicreg, zbontar2021barlow}, or using architecture design \citep{chen2020simsiam, grill2020byol}. Their success is typically measured by the accuracy of a simple classifier (say a shallow network) trained on the representations that they produce, which have progressively improved over the years. Representations learned from these approaches, however, do not emphasize much about the intrinsic structure of the data distribution, and have not demonstrated success for generative purposes.

In parallel, generative methods like GANs \citep{goodfellow2014generative} and VAEs \citep{kingma2013auto} have also been explored for unsupervised learning. Although generative methods have made striking progress in the quality of the sampled or autoencoded data, when compared to the aforementioned discriminative methods, representations learned with these approaches demonstrate inferior performance in classification. %     For example, surveying the latest representative methods from these two categories indicates that there is about a 10\%\by{would be nice to have a citation for where this percentage came from} gap in accuracy on datasets like CIFAR-10 \citep{krizhevsky2014cifar} and CIFAR-100 \citep{krizhevsky2009learning}.

\textbf{Toward A Unified Representation?} The disparity between discriminative and generative approaches in unsupervised learning, contrasted against the fundamental goal of learning representations that are useful across many tasks, leads to a natural question that we investigate in this paper:
\textit{in the unsupervised setting, is it possible to learn a unified representation that is effective for both discriminative and generative purposes? Further, do they mutually benefit each other?}
Concretely, we aim to learn a {\em structured representation} with the following two properties: 
\vspace{-1mm}
\begin{enumerate}
\item The learned representation should be discriminative, such that simple classifiers applied to learned features yield high classification accuracy. % good classification performance.
\vspace{-1mm}
\item The learned representation should be generative, with enough diversity to recover raw inputs, and structure that can be exploited for sampling and generating new images.
\vspace{-1mm}
% \item The learned representation should be diverse enough to replay the training data and has good (linear) structures that can be exploited for effective sampling and generating new images.
% \item The learned representation should be discriminative enough, such that simple classifiers applied to the learned features should yield high classification accuracy. % good classification performance.

    % \item The learned distribution of the representation is disentangled and structured, which means the representation has subspace property can serve for discriminative purpose. 
    % \item The learned distribution of the representation is accessible, where ``accessible'' means we can do meaningful manipulation on it. Moreover, the distribution should be derived directly from the data without specifying a prior. Then we can manipulate the model into generative tasks. 
\end{enumerate}

The fact that human visual memory serves both discriminative tasks (for example, detection and recognition) and generative or predictive tasks (for example, via replay) \citep{Keller2018-ez,Josselyn2020MemoryER,2020Vandeven} indicates that this goal is achievable.
%\by{is there a citation justifying that discrimination \& generation in human visual memory uses a unified representation? for mechanisms for recall vs recognition (slightly different!) it seems like there's evidence to the contrary (see discussion/references on ``dissociable activation'' in nature.com/articles/s41467-018-07830-6)}
% In machine learning, for many pragmatic reasons, we often want to learn a representation that can serve both purposes.
Beyond being possible, these properties are also highly practical -- successfully completing generative tasks like unsupervised conditional image generation \citep{hwang2021stein}, for example, inherently requires that learned features for different classes be both structured for sampling and discriminative for conditioning. On the other hand, the generative property can serve as a natural regularization to avoid representation collapse.
%\ym{I am rewritting the following paragraphs...}

\textbf{Closed-Loop Transcription via a Constrained Maximin Game.} The class of linear discriminative representations (LDRs) has recently been proposed for learning diverse and discriminative features for multi-class (visual) data, via optimization of the {\em rate reduction} objective \citep{chan2021redunet}.
In the supervised setting, these representations have been shown to be be both discriminative and generative if learned in {\em a closed-loop transcription} framework via a maximin game over the rate reduction utility between an encoder and a decoder \citep{dai2022ctrl}.
% All these early works are for the {\em supervised setting} with all classes trained {\em jointly}. 
Beyond the standard joint learning setting, where all classes are sampled uniformly throughout training, the closed-loop framework has also been successfully adapted to the {\em incremental setting} \citep{tong2022incremental}, where the optimal multi-class LDR is learned one class at a time. In the incremental (supervised) learning setting, one solves a {\em constrained maximin problem} over the rate reduction utility which keeps learned memory of old tasks intact (as constraints) while learning new tasks. It has been shown that this new framework can effectively alleviate the catastrophic forgetting suffered by most supervised learning methods. 
% It incrementally learns the optimal LDR for data classes through a {\em constrained minimax game} between discriminator and generator for an intrinsic {\em rate reduction} objective \citep{chan2021redunet}. More specifically, the process of learning a new task while trying to keep the learned LDR (the memory) of old classes intact can be cast as a {\em constrained minimax problem}.

\begin{figure}[t]
\centering
\includegraphics[width=0.9\textwidth]{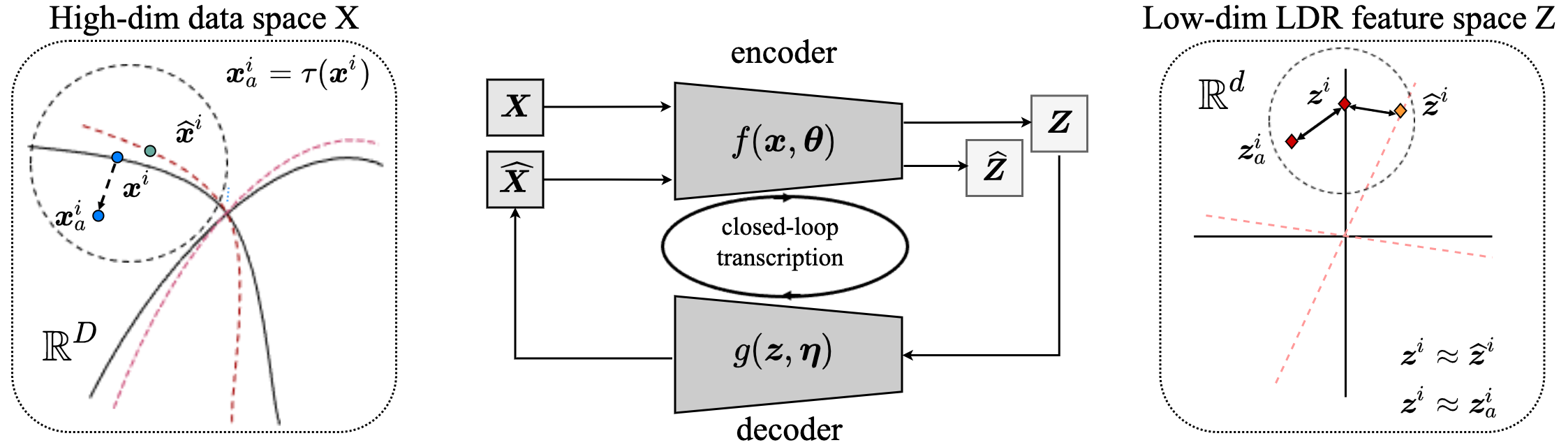}
\caption{\textbf{Overall framework} of closed-loop transcription for unsupervised learning. Two additional constraints are imposed on the Binary-CTRL method proposed in prior work \citep{dai2022ctrl}: 1) self-consistency for sample-wise features $\z^i$ and $\hat{\z}^i$, say $\z^i \approx \hat{\z}^i$; and 2) invariance/similarity among features of augmented samples $\z^i$ and $\z_{a}^i$, say $\z^i \approx \z_{a}^i=f(\tau(\x^i), \theta)$, where $\x^i_a=\tau(\x^i)$ is an augmentation of sample $\x^i$ via some transformation $\tau(\cdot)$.\vspace{-3.5mm}}
\label{fig:framework-uCTRL}
\end{figure}

\textbf{Contributions.} In this work, we show that the closed-loop transcription framework proposed for learning LDRs in the supervised setting \citep{chan2021redunet} can be adapted to a purely unsupervised setting. In the unsupervised setting, we only have to view each sample and its augmentations as a ``new class'' while using the rate reduction objective to ensure that learned features are both {\em invariant}  to augmentation and {\em self-consistent} in generation; this leads to a constrained maximin game that is similar to the one explored for incremental learning~\citep{tong2022incremental}. %, which again can be measured in terms of rate reduct% As we will see, this again leads to a similar type of {constrained minimax game} over the rate reduction utility (see eqn. \eqref{eqn:constrained_maxmin}).
Our overall approach is illustrated in Figure \ref{fig:framework-uCTRL}.

% In this work, we show that this  the multi-class LDR can be learned in an entirely unsupervised way. The only thing one needs to modify is to optimize the rate reduction objective for one sample and its augmented one while trying to enforce the sample wise constraint. It is also a {\em constrained minimax game} (see eqn. \eqref{eqn:constrained_maxmin}). Speak more insight, we formulate the contrastive loss into framework of CTRL as a {\em constrained minimax problem} which can learn a multi-class LDR in an unsupervised way.

%As we will experimentally demonstrate in Section \ref{sec:experiments}, this new framework can indeed efficiently learn a unified representation. Not only can it learn a structured generative model that significantly improves visual quality over the state of the art\pt{I think this may be a little overclaimed}  , but it also effectively bridges the gap between unsupervised discriminative and generative methods for classification tasks. As a result, we believe that the close-loop transcription through the (constrained) minimax game between the encoder and decoder over the rate reduction utility has the potential for offering a truly unifying framework for learning a both discriminative and generative representation, a.k.a. structured memory,  in all settings  (supervised, incremental, and unsupervised).
As we experimentally demonstrate in Section \ref{sec:experiments}, our formulation benefits from the mutual benefits of both discriminative and generative properties. It bridges the gap between two formerly distinct set of methods: by standard metrics and under comparable experimental conditions, it enables classification performance on par with and unsupervised conditional generative quality significantly higher than state-of-the-art techniques. Coupled with evidence from prior work, this suggests that the closed-loop transcription through the (constrained) maximin game between the encoder and decoder has the potential to offer a unifying framework for both discriminative and generative representation learning, across supervised, incremental, and unsupervised settings.

\begin{table}[tbh]
    \centering
    \small
    \setlength{\tabcolsep}{6.5pt}
    \renewcommand{\arraystretch}{1.25}
    \begin{tabular}{l|c|c|c}
    \multirow{1}{*}{Method} & \multicolumn{1}{c|}{Linear Probe} & \multicolumn{1}{c|}{Image Generation} & \multicolumn{1}{c}{UCIG}\\
    \hline
    \hline
    SimCLR \citep{chen2020simple}      &  \compareyes    & \compareno & \compareno \\
    MOCO-V2 \citep{he2020momentum}     &  \compareyes    & \compareno & \compareno \\
    ContraD \citep{jeong2021training}       &  \compareyes    & \compareyes  & \compareno \\

    PATCH-VAE \citep{parmar2021dual}      & \compareyes  & \compareyes & \compareno \\

    CTRL-Binary \citep{dai2022ctrl}       & \compareyes     & \compareyes  & \compareno \\
    
    SLOGAN \citep{hwang2021stein} &  \compareno & \compareyes  & \compareyes    \\
    \ours{} (ours)                  & \compareyes     & \compareyes &  \compareyes \\
    \end{tabular}      
    % \vspace{-0.1in}
    \caption{\small Comparison of the downstream task capabilities of different unsupervised learning methods. UCIG refers to Unsupervised Conditional Image Generation \citep{hwang2021stein}.\vspace{-6mm}}
    \label{tab:compare_method}
\end{table}

\vspace{-2mm}
\section{Related Work}
\vspace{-3mm}
Our work is mostly related to three categories of  unsupervised learning methods: (1) self-supervised learning via discriminative models, (2) self-supervised learning via generative models, and (3) unsupervised conditional image generation. Table \ref{tab:compare_method} compares the capabilities of models learned by various representative unsupervised learning methods. 

\textbf{Self-Supervised Learning for Discriminative Models.}
On the discriminative side, works like SimCLR~\citep{chen2020simple}, MoCo \citep{he2020momentum}, and BYOL \citep{grill2020bootstrap} have recently shown overwhelming effectiveness in learning discriminative representations of data. MoCo \citep{he2020momentum} and SimCLR \citep{chen2020simple} seek to learn features by pulling together features of augmented versions of the same sample while pushing apart features of all other samples, while BYOL \citep{grill2020bootstrap} trains a student network to predict the representation of a teacher network in a contrastive setting. BarlowTwins \citep{zbontar2021barlow} and TCR \citep{li2022neural} learn by regularizing the covariance matrix of the embedding. However, features learned by this class of methods are typically highly compressed, and \textit{not} designed to be used for generative purposes.

\textbf{Self-Supervised Learning with Generative Models.}
On the generative side, the original GAN \citep{goodfellow2014generative} can be viewed as a natural self-supervised learning task. With an additional linear probe, works like DCGAN \citep{radford2015unsupervised} have shown that features in the discriminator can be used for discriminative tasks. To further enhance the features, extensions like BiGAN \citep{donahue2016adversarial} and ALI \citep{dumoulin2016adversarially} introduce a third network into the GAN framework, aimed at learning an inverse mapping for the generator, which when coupled with labeled images can be used to study and supervise semantics in learned representations. Other works like SSGAN \citep{chen2019self}, SSGAN-LA \citep{hou2021self}, and ContraD \citep{jeong2021training} propose to put augmentation tasks into GAN training to facilitate representation learning.
Outside of GANs, variational autoencoders (VAEs) have been adapted to generate more semantically meaningful representations by trading off latent channel capacity and independence constraints with reconstruction accuracy \citep{higgins2016beta}, an idea that has also been incorporated into recognition improvements using patch-level bottlenecks \citep{gupta2020patchvae}, which encourage a VAE to focus on useful patterns in images. By incorporating data-augmentation, VAE is also shown to achieve fair discriminative performance \citep{falcon2021aavae}.
Recently, works like MAE \citep{he2021masked} and CAE \citep{chen2022context} have learned representations by solving masked reconstruction tasks using vision transformers. Autogressive approaches like iGPT \citep{chen2020generative} have also demonstrated decent self-supervised learning performance, which improves further with the incorporation of contrastive learning \citep{kim2021hybridiGPT}.
However, unless supervised, features learned by those previously mentioned methods either do not have strong discriminative performance, or cannot be directly exploited to condition the generative task.

%these methods often struggle to perform tasks like unsupervised conditional image generation.
%\by{this feels a bit vague}\pt{It means to say that general GAN/VAE can hardly do conditional generation when trained in unsupervised setting. To explain it takes a whole paragraph emmm} of GANs, VAEs, and autoencoders, these methods often struggle to perform tasks like unsupervised conditional image generation.

\textbf{Unsupervised Conditional Image Generation (UCIG).} For generative models, we often want to be able to generate images conditioned on a certain class or style, even in a completely unsupervised setting. This requires that the learned representations have structures that correspond to the desired conditioning.
% \ym{Clearly explain what UCIG means and citep the most relevant.} With recent developments in generative models, conditionally generating images using models trained in unsupervised settings has %, which we refer to as unsupervised conditional image generation, has becoming a possible task.
InfoGAN \citep{chen2016infogan} proposes to learn interpretable representations by maximizing the mutual information between the observation and a subset of the latent code. ClusterGAN \citep{mukherjee2019clustergan} assumes a discrete Gaussian prior where discrete variables are defined as a one-hot vector and continuous variables are sampled from Gaussian distribution. Self-Conditioned GAN \citep{liu2020diverse} uses clustering of discriminative features as labels to train. SLOGAN \citep{hwang2021stein} proposes a new conditional contrastive loss (U2C) to learn latent distribution of the data. Note that compared to our work, ClusterGAN and SLOGAN introduce an additional encoder that leads to increased computational complexity. On the VAE side, works like VaDE \citep{jiang2016variational} cluster based on the learned feature of a supervised ResNet. Variational Cluster \citep{prasad2020variational} simultaneously learns a prior that captures the latent distribution of the images and a posterior
to help discriminate between data points in an end-to-end unsupervised setting. In this work, we will see how clusters can be estimated in a principled way in a more unified framework, by optimizing the same type of objective function that we use for learning features.

% \textcolor{purple}{ZL: I feel like autoregressive models probably need to be mentioned too, both here and in intro. iGPT achieves fair self-supervised learning performance, and also can generate samples. Also, should we stress the unique cluster conditioned generation ability early in the paper?}
% \by{+1 for autoregressive, for completeness potentially also flows, diffusion/score-based models?}
% \pt{Good Point! I added some to describe iGPT. For diffusion/score-based, I did some search but didn't find ones that are mostly related to our work. Feel free to add some.}

\vspace{-2mm}
\section{Method}
\vspace{-2mm}
%Our work builds upon the CTRL framework proposed by \citep{dai2022ctrl}. We 
%\pt{Logic: First state unsupervised generative tasks -> Need clean subspace(i.e use CTRL) -> State Method}

\subsection{Preliminaries: Rate Reduction and Closed-Loop Transcription}
\vspace{-2mm}
%\ym{This subsection overlaps with CTRL paper. We should introduce more relevant to our context here, with added information and value. Probably more relevant to introduce the incremental setting... and we can show what we do differently for the unsupervised setting.}

\paragraph{Assumptions on Data.}
Our work, as well as prior work in closed-loop transcription~\citep{dai2022ctrl,tong2022incremental}, considers a set of $N$ images $\X = [\x^1, \x^2, ..., \x^N] \subset \Re^D$ sampled from $k$ classes. Borrowing notation from \citep{yu2020learning}, the membership of the $N$ samples in the $k$ classes is denoted using $k$ diagonal matrices: $\bm{\Pi} = \{\bm{\Pi}_j \in \Re^{N\times N}\}_{j=1}^{k}$, where the diagonal entry $\bm{\Pi}_{j}(i,i)$ of $\bm{\Pi}_{j}$ is the probability of sample $i$ belonging to subset $j$. Let  $\Omega \doteq \{\bm \Pi \mid \sum \bm{\Pi}_j = \bm I, \bm \Pi_j \ge \bm 0.\}$ be the set of all such matrices. WLOG, we may assume that classes are separable, with images for each belonging to a low-dimensional submanifold in the space $\mathbb{R}^D$. % That is, these classes are in theory separable. 

\vspace{-1mm}
\paragraph{Unsupervised Discriminative Autoencoding.}
The goal of transcription is to learn a unified representation, with the structure required to both classify and generate images from these $k$ classes. 
Concretely, this is achieved by learning two continuous mappings: (1) an encoder parametrized by $\theta$: $f(\cdot, \theta): \x \mapsto \z \in \Re^d$ with $d \ll D$ such that all samples are mapped to their features as $\X \xrightarrow{ f(\x, \theta)} \Z$ with $\Z=[\z^1, \z^2, ..., \z^N] \subset \Re^d$, and (2) an inverse map $g(\cdot, \eta): \z \mapsto \hat \x \in \Re^D$ such that $\x$ and $\hat \x = g(f(\x))$ is close. In other words, $\X \xrightarrow{ f(\x, \theta)}  \Z \xrightarrow{ g(\z,\eta)} \hat \X$ forms an autoencoding. 

In this work, we specifically learn this mapping in an {\em entirely unsupervised} fashion, without knowing the ground-truth class labels $\bm \Pi$ at all. As stated in the introduction, a both discriminative and generative representation is difficult to achieve by standard generative methods like VAEs and GANs.
This is one of the motivations for the closed-loop transcription framework (CTRL) proposed by \citep{dai2022ctrl}, which we will generalize to the unsupervised setting. 

\vspace{-1mm}
\paragraph{Maximizing Rate Reduction.} The CTRL framework~\citep{dai2022ctrl} was proposed for the supervised setting, where it aims to map each class onto an independent linear subspace.
% BY: seems equivalent to the next sentence?
% This objective can be conveniently measured by the {\em coding rate reduction} \citep{yu2020learning}. 
As shown in \citep{yu2020learning}, such a linear discriminative representation (LDR) can be achieved by maximizing a coding rate reduction objective, known as {\em the MCR$^2$ principle}: 
\vspace{-2mm}
% %\rc{vspace -2mm here if paper is too long}
% \begin{small}
% \begin{align}
% \max_{\theta} \Delta R(\Z) = \Delta R(\Z_1, \ldots, \Z_k) \doteq  \underbrace{\frac{1}{2}\log\det \big(\I + {\alpha} \Z \Z^{*} \big)}_{R(\Z)} - \sum_{j=1}^{k}\gamma_j\underbrace{\frac{1}{2} \log\det\big(\I + {\alpha_j} \Z_j \Z^{*}_j \big)}_{R(\Z_j)},\nonumber
% \end{align}
% \end{small}where, for a prescribed quantization error $\epsilon$, $ \alpha=\frac{d}{n\epsilon^2},\, \alpha_j=\frac{d}{|\Z_{j}|\epsilon^2}, 
% \, \gamma_j=\frac{|\Z_{j}|}{n}.$
% Or equivalently we write the membership of each datapoint into 
\begin{equation}
     \Delta R\big(\bm Z|\bm \Pi) \doteq \underbrace{\frac{1}{2}\log\det\left(\I + \frac{d}{N\epsilon^{2}}\Z\Z^{\top}\right)}_{
    \large{R(\Z)}} - \underbrace{\sum_{j=1}^{k}
    \frac{\tr(\bm{\Pi}_j)}{2N}\log\det\left(\I + \frac{d}{\tr(\bm{\Pi}_j)\epsilon^{2}}\Z\bm{\Pi}_j\Z^{\top}\right)}_{
    \large{R^c}}.
    %, \quad \mbox{s.t.} \ \|\Z_j(\theta)\|_F^2 = m_j, \, \bm{\Pi} \in {\Omega}.
    \label{eqn:maximal-rate-reduction}
\end{equation}
where each $\bm{\Pi}_j$ encodes the membership of the $N$ samples described before. As discussed in \citep{chan2021redunet}, the first term $R(\Z)$ measures the total rate (volume) of all features whereas the second term $R^c$ measures the average rate (volume) of the $k$ components.
Our work adapts this formula to design meaningful objectives in the unsupervised setting.

%in the $k$ classes: the diagonal entry $\bm{\Pi}_{j}(i,i)$ of $\bm{\Pi}_{j}$ is the probability of sample $i$ belonging to subset $j$. $\Omega \doteq \{\bm \Pi \mid \sum \bm{\Pi}_j = \bm I, \bm \Pi_j \ge \bm 0.\}$ 

%The first term  is the coding rate of the whole feature set $\Z$ (coded as a Gaussian source) with a prescribed precision $\epsilon$; the second term is the average coding rate of the $k$ subsets of features $\Z_j = f(\X_j)$ (each coded as a Gaussian). 

% As it has been shown by \citep{yu2020learning}, maximizing the difference between the two terms will expand the whole feature set while compressing and linearizing features of each of the $k$ classes. If the mapping $f$ maximizes the rate reduction, it maps the features  of different classes into independent (orthogonal) subspaces in $\Re^d$. Figure \ref{fig:MCR2} illustrates a simple example of data with $k=2$ classes (on two submanifolds) mapped to two incoherent subspaces (solid black lines). Notice that, compared to AE \eqref{eqn:autoencoding} and GAN \eqref{eqn:GAN}, the above mapping \eqref{eqn:LDR} is only one-sided: from the data $\X$ to the feature $\Z$. In this work, we will see how to use the rate reduction metric to establish the inverse mapping from the feature $\Z$ back to the data $\X$, while still preserving the subspace structures in the feature space.

%As noted in \citep{yu2020learning}, maximizing the rate reduction promote learned features that span the entire feature space. 
 
\vspace{-1mm}
\paragraph{Closed-Loop Transcription.} To learn the autoencoding $\X \xrightarrow{ f(\x, \theta)}  \Z \xrightarrow{ g(\z,\eta)} \hat \X$, a fundamental question is how we measure the difference between $\X$
and the regenerated $\hat \X = g(f(\X))$. It is typically very difficult to put a proper distance measure in the image space \citep{image-similarity}. To bypass this difficulty, the {\em closed-loop transcription} framework~\citep{dai2022ctrl} proposes to measure the difference between $\X$ and $\hat \X$ through the  difference between their  features $\Z$ and $\hat \Z$ mapped through the same encoder:
% it is shown that rate reduction can be extended nicely to generative purposes and one of its variant CTRL-Binary functions under fully unsupervised setting(i.e label-unaware).
% Now considering learning the entire data $\X$ via a set of autoencoder that consists of an encoder $f(\cdot, \theta)$, parameterized by $\theta$, that  maps the data $\x \in \Re^D$ continuously to a compact feature $\z$ in a much lower-dimensional space $\Re^d$, and a decoder $g(\cdot, \eta)$, parameterized by $\eta$, that maps a feature $\z$ back to the original data space $\Re^D$:
% \begin{equation}
% f(\cdot, \theta): \x \mapsto \z \in \Re^d; \quad g(\cdot, \eta): \z \mapsto \hat \x \in \Re^D. \vspace{-1mm}
% \end{equation}
% we have two sets of features: the features $\Z = f(\X, \theta)$ for the original data $\X$ and those $\hat \Z = f(\hat \X(\theta,\eta), \theta) $ for replayed data $\hat \X = g(\Z(\theta), \eta)$, through a ``closed-loop'' transcription:
\vspace{-2mm}
\begin{equation}
\X \xrightarrow{\hspace{2mm} f(\x, \theta)\hspace{2mm}}  \Z \xrightarrow{\hspace{2mm} g(\z,\eta) \hspace{2mm}} \hat \X \xrightarrow{\hspace{2mm} f(\x, \theta)\hspace{2mm}} \ \hat \Z.
\end{equation}
The difference can be measured by the rate reduction between $\Z$ and $\hat{\Z}$, a special case of \eqref{eqn:maximal-rate-reduction} with $k=2$ classes:
\vspace{-2mm}
\begin{equation}
\Delta R\big(\Z, \hat{\Z}\big) \doteq R\big(\Z \cup \hat{\Z}\big) - \frac{1}{2} \big( R\big(\Z) + R\big(\hat \Z)\big).
\end{equation}
Such a $\Delta R$ is a principled distance between subspace-like Gaussian ensembles, with the property that $\Delta R(\Z, \hat{\Z}) = 0$ iff $\mbox{Cov}(\Z) = \mbox{Cov}(\hat{\Z})$ \citep{ma2007segmentation}. %Hence, learning and generating the distribution can be viewed as a minimax game between the encoder $f(\cdot , \theta)$ and $g(\cdot, \eta)$: while the encoder tries the maximize the rate reduction between the original $\Z$ and reconstructed $\hat{Z}$, the decoder tries the minimize it. 

As shown in \citep{dai2022ctrl}, applying this measure in the closed-loop CTRL formulation can already learn a decent autoencoding, even without class information.
This is known as the  CTRL-Binary program:\vspace{-1mm}
\begin{align}
      \max_\theta \min_\eta \quad \Delta R(\Z, \hat{\Z}) 
 \label{eqn:CTRL-Binary}\vspace{-2mm}
\end{align}
However, note that \eqref{eqn:CTRL-Binary} is practically limited because it only aligns the dataset $\X$ and the regenerated $\hat \X$ at the distribution level. %, and neglects self-consistency (between $\x$ and $\hat \x$) for individual samples. 
There is no guarantee that for each sample $\x$ would be close to the decoded $\hat \x = g(f(\x))$. For example, \citep{dai2022ctrl} shows that a car sample can be decoded into a horse; the so obtained (autoencoding) representations are not sample-wise \textit{self-consistent}!
%Hence, we cannot naively adopt the CTRL framework for learning without class information. %, and instead, we need to extend it to the unsupervised or self-supervised setting in a principled way.

% t is not suitable to naively apply for the case of unsupervised learning, as the representation for an individual sample is unknown and very unstable for future tasks like unsupervised conditional image generation. To address these problems, we introduce a few amendments in the next subsection. 

\vspace{-4mm}

%It is therefore not suitable to naively apply for the case of incremental learning, as the number of classes increases within a fixed feature space.\footnote{As the number of classes is initially small in the incremental setting, if the dimension of the feature space $d$ is high, maximizing the rate reduction may over-estimate the dimension of each class.} The closed-loop transcription framework introduced by \citep{dai2021closedloop} suggests resolving such difficulties by suggesting learning the encoder $f(\cdot, \theta)$ and decoder $g(\cdot, \eta)$ together as a minimax game: while the encoder tries to maximize the rate reduction objective, the decoder should minimize it instead. That is, the decoder $g$ minimizes resources (measured by the coding rate) needed for the replayed data for each class $\hat \X_j = g(\Z_j, \eta)$, decoded from the learned features $\Z_j = f(\X_j,\theta)$, to emulate the original data $\X_j$ well enough.  As it is typically difficult to directly measure the similarity between $\X_j$ and $\hat \X_j$, \citep{dai2021closedloop} proposes measuring this similarity with the {\em rate reduction} of their corresponding features $\Z_j$ and $\hat \Z_j = f(\hat \X_j(\theta,\eta),\theta)$:

\subsection{Sample-Wise Constraints for Unsupervised Transcription} 
\label{sec:constraints}
\vspace{-2mm}
%\ym{The entire Section 3.2 needs to be rewritten completely now...}

To improve discriminative and generative properties of representations learned in the unsupervised setting, we propose two additional mechanisms for the above CTRL-Binary maximin game \eqref{eqn:CTRL-Binary}. % based on two additional fundamental mechanisms.
For simplicity and uniformity, here these will be formulated as equality constraints over rate reduction measures, but in practice they can be enforced softly during optimization.
\vspace{-2mm}
\paragraph{Sample-wise Self-Consistency via Closed-Loop Transcription.} 
First, to address the issue that CTRL-Binary does not learn a sample-wise consistent autoencoding, we need to promote $\hat \x$ to be close to $\x$ for each sample. In the CTRL framework, this can be achieved by enforcing that their corresponding features $\z=f(\x)$ and $\hat \z = f(\hat \x)$ are the same or close. 
% a cat\by{is this cat or car? earlier paragraph says car} could be decoded as a horse in CTRL-Binary, we promote self-consistency within the closed-loop for each sample; that is, a cat should be decoded back to a cat. An intuitive approach is directly bridging the distance between feature $\z$ of a sample and feature $\hat{\z}$ of the reconstructed sample. 
%Recent work \citep{ayub2021eec} has further suggested that if we want the visual details of $\hat \x$ to be close to $\x$, it is better to enforce some of the information-rich features in an intermediate convolution layer, in our case the second-to-last one, to be the same or close. We apply this idea in our framework. Let $\z_{conv}$ denote the feature of $\x$ from the second-to-last convolution layer of the encoder network $f(\cdot, \theta)$; similarly $\hat{\z}_{conv}$ is the feature of the decoded image $\hat{\x} = g(\z)$ from the second-to-last convolution layer. % it is more effective to minimize the distance between the internal convolutional feature within a neural network, typically from the second-to-last convolutional layer. To do so, we re-define the encoder to give both the representation and convolutional feature in the middle $f:(\cdot, \theta):\x \mapsto (\z, \z_{conv})$ where $\z$ is the final feature and $\z_{conv}$ is the convolutional feature from the second-to-last convolutional layer of the encoder network.
To promote sample-wise self-consistency, where $\hat{\x} = g(f(\x))$ is close to $\x$ , we want the distance between $\z$ and $\hat{\z}$ to be zero or small, for all $N$ samples.
This can be formulated using rate reduction; note that this again avoids measuring differences in the image space:
%\pt{Shorten down--introduce the ayub2021eec work, define $z_{conv}$ as further internal layer, then introduce the following equation for loss.}
% sample-wise alignment also using the rate reduction objective. An intuitive approach will be directly bridging the distance between $\z$ and reconstructed $\hat{\z}$.\ym{here is $\hat{\z}$, and it is switched to $\z$ below... without any explanation...} However, It is shown in EEC \citep{ayub2021eec} that it is more effective to minimize the distance between the convolutional feature from the last convolutional layer of the encoder network instead of minimzing distance between $\z$ and $\hat{\z}$. To do so, we re-define the encoder to be $f:(\cdot, \theta):\x \mapsto (\z, \z_{conv})$ where $\z_{conv}$ is the  convolutional feature from the last convolutional layer of the encoder network.\ym{I have no idea what this means: is $\z_{conv}$  with a hat or not!? if it is with a hat, it also depends on the closed-loop! Also why you want $\z$ and this $\z_{conv}$ or $\hat{\z}_{conv}$ to be close? Is $\z$ also the last second layer? Why not the last layer? No explanation??? Guys, this is not clear at all. }
% \vspace{-1mm}
\begin{align}
\sum_{i\in N} \Delta R(\z^i,\hat{\z}^i) = 0.
\label{eqn:sample-self-consistency}\vspace{-2mm}
\end{align}

\paragraph{Self-Supervision via Compressing Augmented Samples.} 
Since we do not know any class label information between samples in the unsupervised settings, the best we can do is to view every sample and its augmentations (say via translation, rotation,  occlusion etc) as one ``class'' --- a basic idea behind almost all self-supervised learning methods. In the rate reduction framework, it is natural to compress the features of each sample and its augmentations. In this work, we adopt the standard transformations in SimCLR \citep{chen2020simple} and denote such a transformation as $\tau$. We denote each augmented sample $\x_a = \tau(\x)$, and its corresponding feature as $\z_a = f(\x_a, \theta) $. For discriminative purposes, we hope the classifier is {\em invariant} to such transformations. Hence it is natural to enforce that the features $\z_a$ of all augmentations are the same as that $\z$ of the original sample $\x$. This is equivalent to requiring the distance between $\z$  and  $\z_a$, measured in terms of rate reduction again, to be zero (or small) for all $N$ samples: 
%\vspace{-1mm}
\begin{align}
\sum_{i\in N} \Delta R(\z^i,\z_{a}^i) = 0.
\label{eqn:sample-compression}\vspace{-3mm}
\end{align}

\subsection{Unsupervised Representation Learning via Closed-Loop Transcription} 
\vspace{-2mm}
% \textbf{Unsupervised Closed-Loop Data Transcription.}
So far, we know the CTRL-Binary objective $\Delta R(\Z, \hat{\Z})$ in \eqref{eqn:CTRL-Binary} helps align the distributions while sample-wise self-consistency \eqref{eqn:sample-self-consistency} and sample-wise augmentation \eqref{eqn:sample-compression} help align and compress features associated with each sample. Besides consistency, we also want learned representations are maximally discriminative for different samples (here viewed as different ``classes''). Notice that the rate distortion term $R(\Z)$ measures the coding rate (hence volume) of all features. It has been observed in \citep{li2022neural} that by maximizing this term, learned features expand and hence become more discriminative. 

\paragraph{Unsupervised CTRL.} Putting these elements together, we propose to learn a representation via the following constrained maximin program, which we refer to as {\em unsupervised CTRL} (\ours{}):%\ym{Switching to maximin instead of minimax to be more theoretically sound...}
% In this work, we present and study {\em unsupervised CTRL (u-CTRL)}, which we formulate by combining CTRL-Binary, our two proposed constraints, and the additional expansion term.
% Concretely, u-CTRL aims to solve the following constrained minimax program:
%By incorporating this expansion term and the constraints, the minimax program \eqref{eqn:CTRL-Binary} can be adapted into a constrained minimax game, which we refer to as unsupervised CTRL (u-CTRL):
\begin{align}
      \max_\theta \min_\eta  \quad & R(\Z) + \Delta R(\Z, \hat{\Z}) \label{eqn:constrained_maxmin}\\
 \mbox{subject to} \quad & \sum_{i\in N} \Delta R(\z^i, \hat{\z}^i) = 0, \;\; \mbox{and} \;\; \sum_{i\in N} \Delta R(\z^i, \z_{a}^i) = 0. \nonumber
\vspace{-2mm}
\end{align}
In practice, the above program can be optimized by alternating maximization and minimization between the encoder $f(\cdot,\theta)$ and the decoder $g(\cdot,\eta)$. We adopt the following optimization strategy that works well in practice, which is used for all subsequent experiments on real image datasets:
\vspace{-1mm}
\begin{align}
  &  \max_{\theta}\; R(\Z) + \Delta{R(\Z, \hat{\Z})-\lambda_{1}\sum_{i\in N} \Delta R(\z^i, \z_{a}^i)} -\lambda_{2}\sum_{i\in N} \Delta R(\z^i, \hat{\z}^i) \label{eqn:constrained_max}; \\
% \end{align}
% \begin{align}
   & \min_{\eta}\; R(\Z) + \Delta{R(\Z, \hat{\Z})+\lambda_{1}\sum_{i\in N} \Delta R(\z^i, \z_{a}^i)+ \lambda_{2}\sum_{i\in N} \Delta R(\z^i, \hat{\z}^i)} \label{eqn:constrained_min}, 
\end{align}
where the constraints $\sum_{i\in N} \Delta R(\z^i, \hat{\z}^i) = 0$ and $\sum_{i\in N} \Delta R(\z^i, \z_{a}^i) = 0$ in \eqref{eqn:constrained_maxmin} have been converted (and relaxed) to Lagrangian terms with corresponding coefficients $\lambda_{1}$ and  $\lambda_{2}$.\footnote{Notice that computing the rate reduction terms $\Delta R$ for all samples or a batch of samples requires computing the expensive $\log\det$ of large matrices. In practice, from the geometric meaning of $\Delta R$ for two vectors, $\Delta R$ can be approximated with an $\ell^2$ norm or the cosine distance between two vectors.}

%\pt{Note here since our actual implementation for augmentation and nst is l2 to save computational resource. Even though Delta R and l2 are essentially the same, how should we note it here?} \ym{Show mathematically that $\Delta R$ can be approximated well by l2, between two vectors.}

\begin{figure}[t]
     \footnotesize
     \centering
     \subfigure[CIFAR-10]{
         \includegraphics[width=0.25\textwidth]{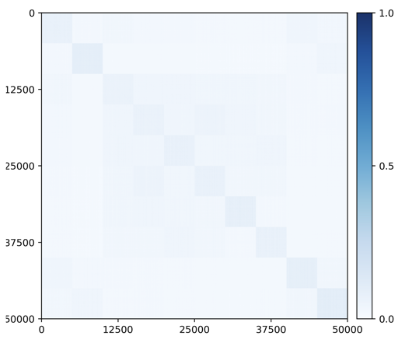}
     }
     \subfigure[CIFAR-100]{
         \includegraphics[width=0.305\textwidth]{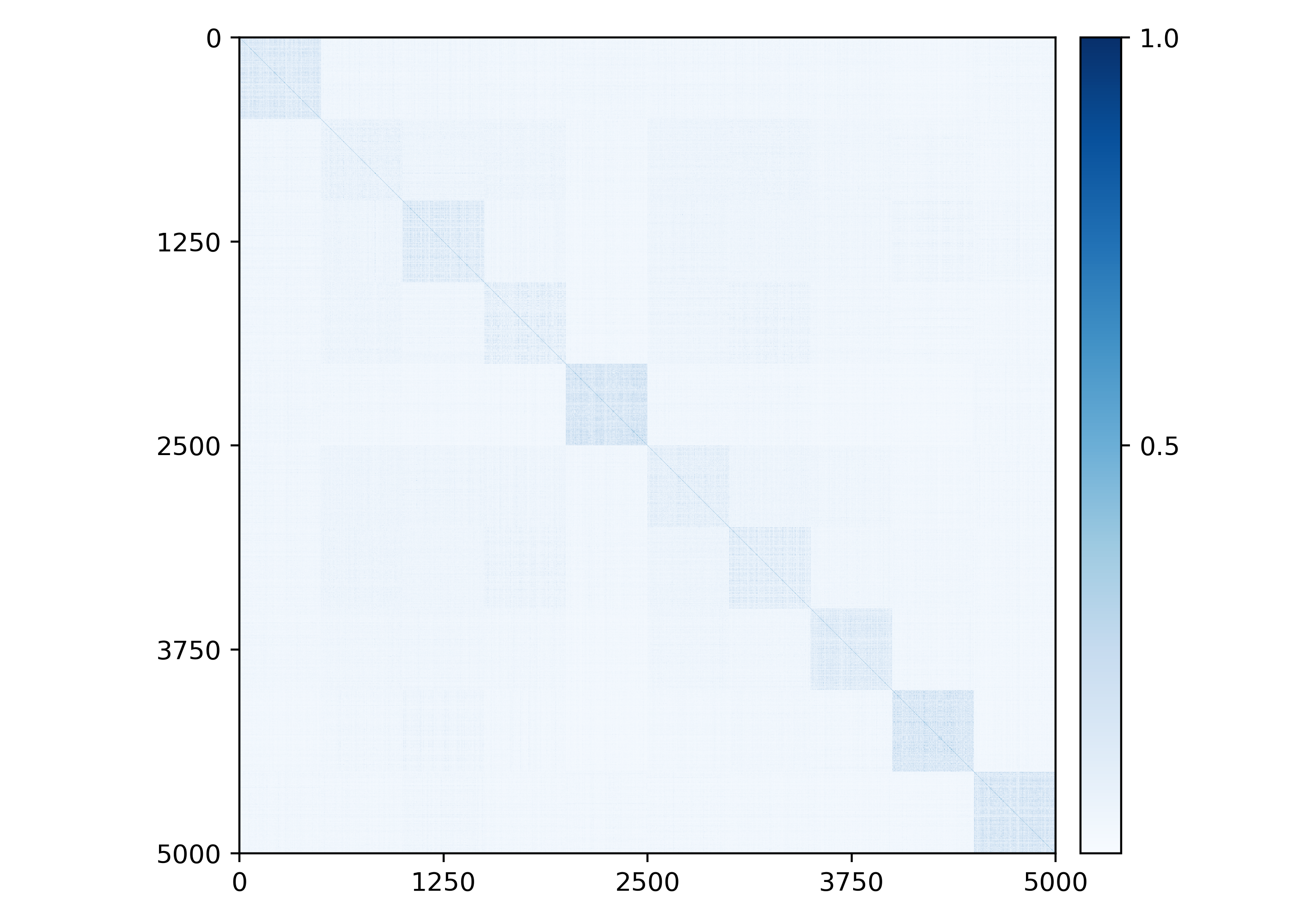}
     }
     \subfigure[Tiny ImageNet]{
         \includegraphics[width=0.305\textwidth]{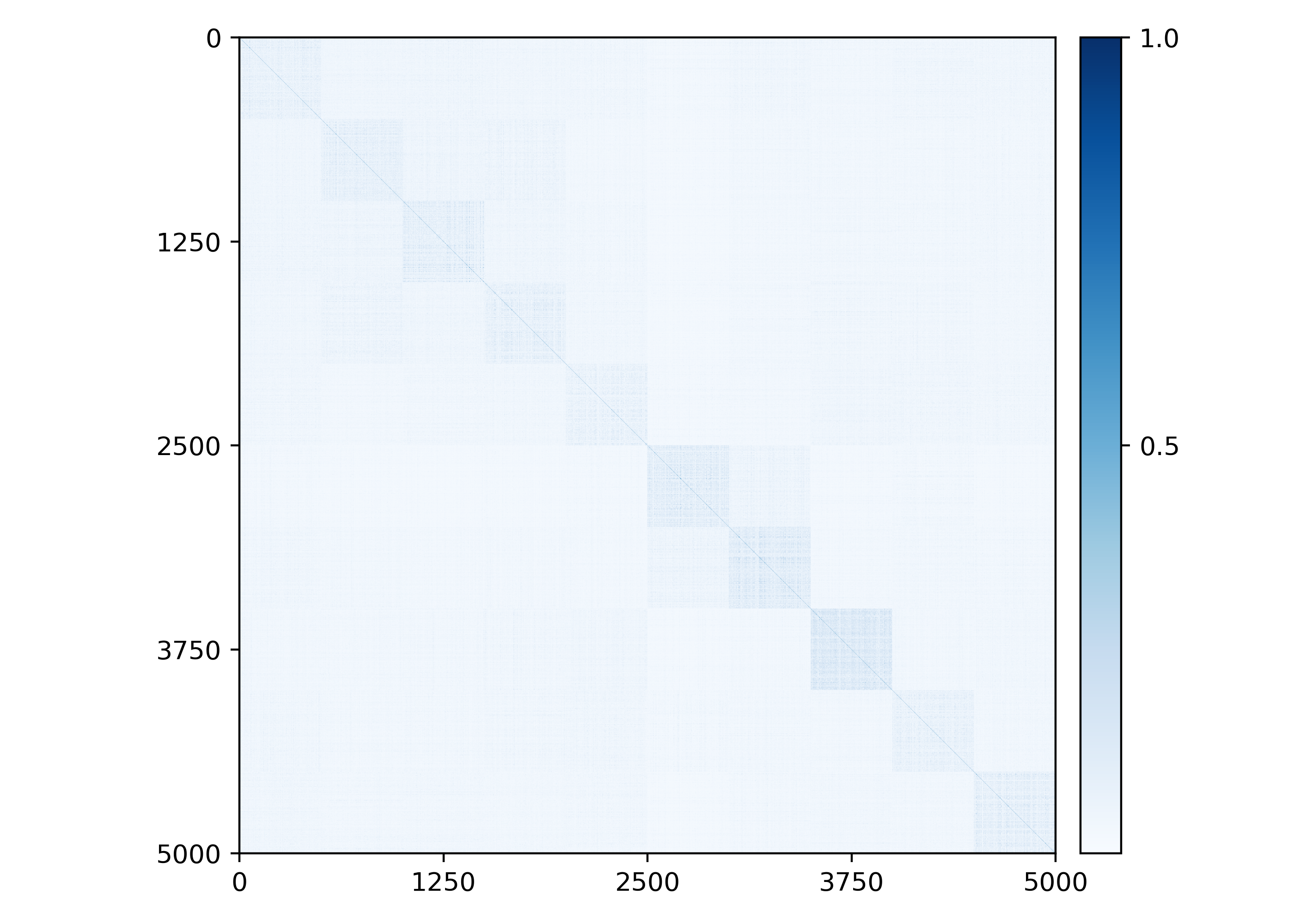}
     }
    \vspace{-0.15in}
    \caption{\small Emergence of block-diagonal structures of $|\Z^\top \Z|$ in the feature space for CIFAR-10 (left), 10 random classes from CIFAR-100 (middle), and 10 random classes from Tiny ImageNet (right). \vspace{-6mm}}
    \label{fig:heatmap_z}
\end{figure}

\vspace{-2mm}
\paragraph{Unsupervised Conditional Image Generation via Rate Reduction.}
The above representation is learned without class information. In order to facilitate discriminative or generative tasks, it must be highly structured.
% Rather surprisingly, that seems to be the case here.
As we will see via experiments, specific and unique structure indeed emerges naturally in the representations learned using \ours{}: globally, features of images in the same class tend to be clustered well together and separated from other classes (Figure \ref{fig:heatmap_z}); locally, features around individual samples exhibit approximately piecewise linear low-dimensional structures (Figure \ref{fig:tsne}). 

% As we will see in the experimental section, a simple classifier already gives good performance (see Section  \ref{sec:feature-classification}).
The highly-structured feature distribution also suggests that the learned representation can be very useful for generative purposes. For example, we can organize the sample features into meaningful clusters, and model them with low-dimensional (Gaussian) distributions or subspaces. By sampling from these compact models, we can conditionally regenerate meaningful samples from computed clusters. This is known as {\em unsupervised conditional image generation} \citep{hwang2021stein}.  % Beyond learning a discriminative representation, we consider the task of unsupervised organization of samples into meaningful clusters and conditionally generating them~\citep{hwang2021stein}. %y generate them, we call unsupervised conditional image generation~\citep{hwang2021stein}.

To cluster features, we exploit the fact that the rate reduction framework \eqref{eqn:maximal-rate-reduction} is inspired by unsupervised clustering via compression \citep{ma2007segmentation}, which provides a principled way to find the membership $\bm \Pi$.
Concretely, we maximize the same rate reduction objective \eqref{eqn:maximal-rate-reduction} over $\bm \Pi$, but fix the learned representation $\Z$ instead. We simply view the membership $\bm \Pi$ as a nonlinear function of the features $\Z$, say $h_{\bm \pi}(\cdot,\xi):\Z \mapsto \bm \Pi$ with parameters $\xi$. In practice, we model this function with a simple neural network, such as an MLP head right after the output feature $\z$. 
% we here introduce a very principled way by predicting the membership $\Pi$ using rate reduction. 
% For any fixed learned representation $\z^j = f(\z^j, \theta)$ of datapoint $\x^j$, we can assign its membership $\bm \Pi^j$ by a function $f_{\bm \pi}(\cdot,\xi):\Z \mapsto \bm \Pi$. f can be implemented as simple as a MLP head. %Conditionally generating datapoints $\X$ of k \ym{why so sloppy??} classes can be solved by: \ym{I have no idea what this sentence mean... does every reader needs to know what ``conditional generating datapoints'' means? }
To estimate a ``pseudo'' membership $\hat{\bm \Pi}$ of the samples, we solve the following optimization problem over $\bm \Pi$:
\vspace{-2mm}
\begin{align}
    \hat{\bm \Pi} = \arg\max_{\xi} \Delta R(\Z| \bm \Pi(\xi)).
\label{eqn:cluster_mcr}\vspace{-2mm}
\end{align}
Experiments in Section \ref{sec:conditional-generation} demonstrate that conditional image generation from clusters produced in this manner result in high-quality images that are highly similar in style.

% %\ym{what it is? citation?} is considered a harder task as it requires the model to be both discriminative and easily-accessible.\ym{what does that mean precisely?} In this section, we will show that such task\ym{??} can be easily solved by maximizing the rate reduction.\ym{didn't you just do that? why maximizing again? what is the difference now?} After learning the representation through the constrained minmax\ref{eqn:constrained_maxmin},\ym{??}  we fix the learned representation. For any fixed learned representation $\z^j = f(\z^j, \theta)$ of datapoint $\x^j$, we can assign its membership $\bm \Pi^j$ by a function $f_{\bm \pi}(\cdot,\xi):\Z \mapsto \bm \Pi$. Such a function can be paramerized\ym{??} by a simple neural network, lets call it $f_{\bm \pi}$, which in practice is a simple mlp head. Conditionally generating datapoints $\X$ of k \ym{why so sloppy??} classes can be solved by: \ym{I have no idea what this sentence mean... does every reader needs to know what ``conditional generating datapoints'' means? }
% \begin{align}
%     \max_{\xi} \Delta R(\Z| \bm \Pi(\xi)) 
% \vspace{-3mm}
% \end{align}
% Experiments in the later section will demonstrate that, when a discriminative representaion is learned, classes can be assigned accurately through this very principled way.   

%\textcolor{blue}{ZL: Should we point out this clustering is a subspace clustering process using rate reduction objective, as the ma 2007 paper has done?}

\vspace{-2mm}
\section{Experiments}
\label{sec:experiments}
\vspace{-2mm}
We now evaluate the performance of the proposed U-CTRL framework and compare it with representative unsupervised generative and discriminative methods. The first set of experiments (Section \ref{sec:feature-classification} show that despite being a generative method in nature, U-CTRL can learn discriminative representations competitive with state-of-the-art discriminative methods. The second set (Section \ref{sec:conditional-generation}) show that the learned generative representation can significantly boost the performance of unsupervised conditional image generation.
Finally, the third set (Section \ref{sec:exp_benefit}) study how the advantages that generative represeentations have over discriminative ones.%unsupervised learning may benefit from generative representations comparing to purely discriminative ones.

We conduct experiments on the following datasets: CIFAR-10 \citep{krizhevsky2014cifar}, CIFAR-100 \citep{krizhevsky2009learning}, and Tiny ImageNet \citep{deng2009imagenet}. Standard augmentations for self-supervised learning are used across all datasets \citep{chen2020simple}.

We design all experiments to ensure that comparisons against \ours{} are fair.
For all methods that we compare against, we ensure that experiments are conducted with similar model sizes. If code for similar size structure can not be found, we uniformly use ResNet-18 to reproduce results for baselines, which is larger than the network used by our method.
Details about network architectures and the experimental setting are given in Appendix \ref{sec:appendix_networkarch}. All methods have runned 400 epochs or equivalent iterations (because generative models often count in iteration).

\begin{table}[t]
    \centering
    \small
    \setlength{\tabcolsep}{6.5pt}
    \renewcommand{\arraystretch}{1.25}
    \begin{tabular}{l|c|c|c}
    \multirow{2}{*}{Method} & \multicolumn{1}{c|}{CIFAR-10} & \multicolumn{1}{c|}{CIFAR-100} & \multicolumn{1}{c}{Tiny-ImageNet}\\
    \cline{2-4}
     ~  & Accuracy  & Accuracy & Accuracy \\
    \hline
    \hline
    \textit{GAN based methods}   &  ~     & ~    & ~ \\
    SSGAN-LA\citep{hou2021self}      &  0.803    & 0.543  & 0.344   \\
    DAGAN+\citep{antoniou2017data}        &  0.772    & 0.519  & 0.224   \\
    ContraD\citep{jeong2021training}       &  0.852    & 0.514  & -   \\
    \hline
    \textit{VAE based methods}    & ~       & ~    & ~  \\
    
    PATCH-VAE \citep{parmar2021dual}      & 0.471  & 0.325 & -  \\
    $\beta$-VAE \citep{higgins2016beta}      & 0.531        & 0.315 & -  \\
    \hline
    \textit{CTRL based methods}   &  ~     & ~    & ~ \\

    CTRL-Binary\citep{dai2022ctrl}                 & 0.599     & - & -  \\
    \ours{} (ours)                  & \textbf{0.874}     & \textbf{0.552} &  \textbf{0.360} \\
    \end{tabular}      
    % \vspace{-0.1in}
    \caption{\small Comparison of classification accuracy on CIFAR-10, CIFAR-100, and Tiny-ImageNet with other generative self-supervised learning methods. U-CTRL is clearly better. \vspace{-4mm}}
    \label{tab:comparison_linearProbe_generative}
\end{table}

\begin{table}[t]
    \centering
    \small
    \setlength{\tabcolsep}{6.5pt}
    \renewcommand{\arraystretch}{1.25}
    \begin{tabular}{l|c|c|c}
    \multirow{2}{*}{Method} & \multicolumn{1}{c|}{CIFAR-10} & \multicolumn{1}{c|}{CIFAR-100} & \multicolumn{1}{c}{Tiny-ImageNet}\\
    \cline{2-4}
     ~  & Accuracy  & Accuracy & Accuracy \\
    \hline
    \hline
  
    SIMCLR      &  0.869    & 0.545  & 0.359   \\
    MoCoV2        &  0.872  & \textbf{0.589}  & 0.365   \\
    
    BYOL      & \textbf{0.883}  & 0.581 & \textbf{0.371}  \\
  
    \ours{} (ours)                  & 0.874     & 0.552 &  0.360 \\
    \end{tabular}      
    
    \caption{\small Comparison of classification accuracy on CIFAR-10, CIFAR-100, and Tiny-ImageNet with purely discriminative self-supervised learning methods. U-CTRL is on par with these non-generative methods.\vspace{-4mm}}
    \label{tab:comparison_linearProbe_discriminative}
    \vspace{-0.15in}
\end{table}

\vspace{-3mm}
\subsection{Discriminative Quality of Learned Representations}\label{sec:feature-classification}
\vspace{-2mm}
To evaluate the discriminative quality of the learned representations, we follow the standard practice of evaluating the accuracy of a simple linear classifier trained on the learned representation. Table \ref{tab:comparison_linearProbe_generative} compares our method against SOTA generative self-supervised learning methods, and Table \ref{tab:comparison_linearProbe_discriminative} compares our method against SOTA discriminative self-supervised methods.
Experimental and training details are given in Appendix \ref{sec:appendix_networkarch}.
%For all experiment, we strictly enforce that the encoder is no larger than resnet-18 for fair comparison. The accuracy are calculated by fixing the encoder and linear probe on the encoded representation and training details is attached to the appendix. 

\vspace{-1mm}
\textbf{Quantitative Comparisons of Classification Performance.}  
From Table \ref{tab:comparison_linearProbe_generative}, we observe that on all chosen datasets, our method achieves substantial improvements compared to existing generative self-supervised learning methods. This includes more complex datasets like CIFAR-100 and Tiny-ImageNet, where we surpass the current SOTA models.  %even when our chosen network are smaller than some of others! 
From Table \ref{tab:comparison_linearProbe_discriminative}, our method achieves similar performance compared to SOTA discriminative self-supervised models.
These results echo our goal of seeking a more unifed generative and discriminative representations: despite resembling a generative method architecturally, our method still produces highly discriminative representations. % to bridge the performance gap between generative self-supervised learning methods and discriminative  self-supervised learning methods. %\pt{Discuss what message do we want to convey here?} 
In addition, these results lead us to ask a fundamental question: when is incorporating both discriminative and generative properties a whole greater than the sum of its parts, particularly outside of the context of computational efficiency? We provide preliminary answers in Section~\ref{sec:exp_benefit}.

\begin{figure}[h]
     \footnotesize
     \centering
     \subfigure[CIFAR-10 $\X$]{
         \includegraphics[width=0.215\textwidth]{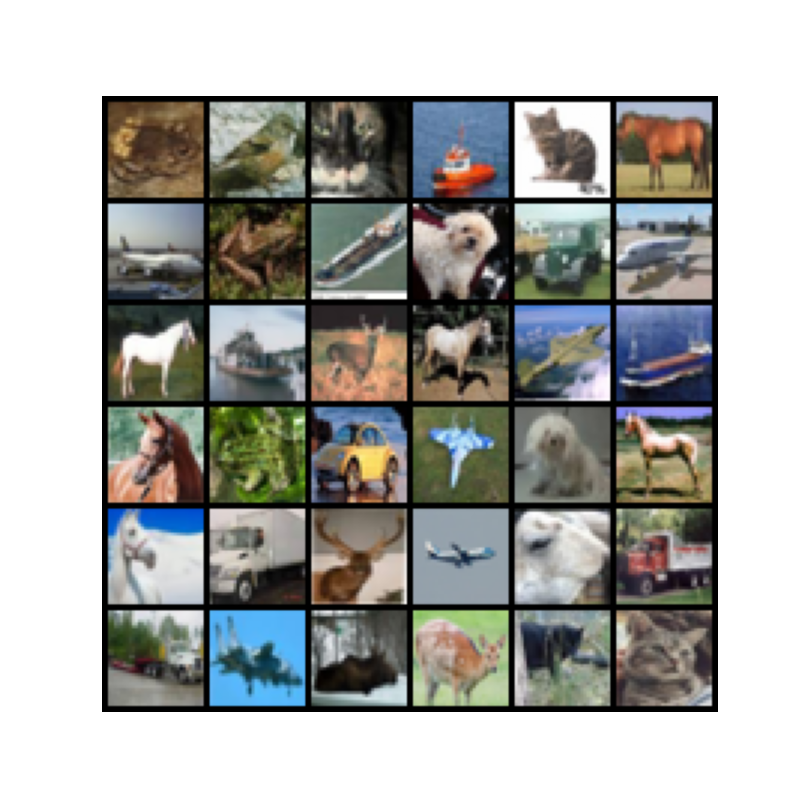}
     }
     \subfigure[CIFAR-10 $\hat{\X}$]{
         \includegraphics[width=0.215\textwidth]{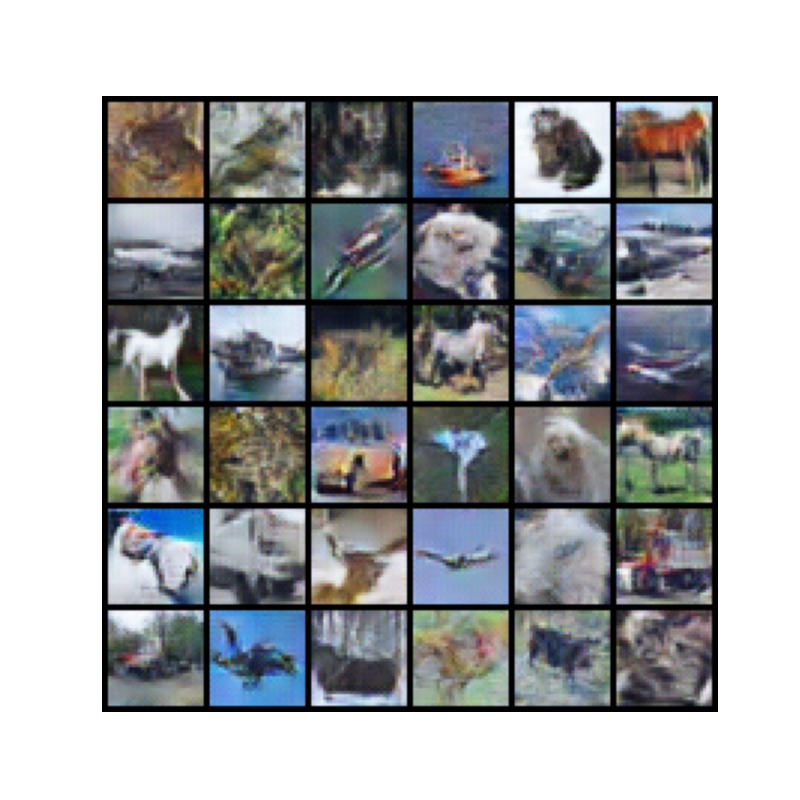}
     }
    \vspace{-0.25in}
    \caption{\small Sample-wise self-consistency: visualization of images $\X$ and reconstructed $\hat{\X}$ on CIFAR-10 dataset. \vspace{-10mm}}
    \label{fig:vis_recon}
\end{figure}

\vspace{-1mm}
\textbf{Qualitative Visualization of Learned Representations.} To explain the classification performance of our method, we visualize the incoherence between features learned for the training datasets. Figure \ref{fig:heatmap_z} shows cosine similarity heatmaps between the learned features, organized by ground-truth class labels. A block-diagonal pattern emerges automatically from \ours{} training for all three datasets, similar to those observed in features learned in a supervised setting \citep{dai2022ctrl}. In this case, however, these blocks emerge and correspond with classes labels despite the absence of any supervision at all.

%\begin{table}[t]
%    \centering
%    \small
%    \setlength{\tabcolsep}{6.5pt}
%    \renewcommand{\arraystretch}{1.25}
%    \begin{tabular}{l|cc|cc|cc}
%    \multirow{2}{*}{Method} & \multicolumn{2}{c|}{CIFAR-10} & \multicolumn{2}{c|}{CIFAR-100} & \multicolumn{2}{c}{ImageNet}\\
%    \cline{2-7}
%     ~  & FID & IS  & FID & IS  & FID & IS \\
%    \hline
%    \hline
%    \textit{GAN based methods}            & ~    & ~    & ~  &  ~   & ~    & ~ \\
%    Self-Conditioned GAN \citep{}& 18.0    & 7.7  & -    & -    & -   &  -   \\
%    SLOGAN \citep{}&  20.6    & -   &  -   & -    & - & -\\
%    SSGAN-LA \pt{need to run} & ~    & ~    & ~  &  ~   & ~    & ~  \\
%    
%    
%    \hline
%    \textit{VAE based methods}      & ~    & ~    & ~   &  ~   & ~    & ~\\
%    k-DAE    \citep{}        & -    & -    & -  &  ~   & ~    & ~ \\
%    LT-VAE    \citep{}        & -    & -    & -  &  ~   & ~    & ~ \\
%    PATCH-VAE \citep{parmar2021dual}       & -    & -    & -  &  ~   & ~    & ~ \\%
%    \hline
%    \textit{CTRL based methods}       & ~    & ~    & ~   &  ~   & ~    & ~\\
%    CTRL-Binary                     & 19.6   & 8.1   & -  &  -   & -    & -\\
%    \ours{}(ours)                     & 17.4   & 8.1   & -  &  -   & -    & -\\
%    \end{tabular}      
%    % \vspace{-0.1in}
%    \caption{\small Comparison on CIFAR-10, STL-10, and ImageNet.}
%    \label{tab:comparsion_quality}
%\end{table}

\vspace{-3mm}
    \subsection{Improved Unsupervised Conditional Generation Quality}\label{sec:conditional-generation}
\vspace{-2mm}
%To successfully perform the task of unsupervised conditional image generation, the following three qualities are the most decisive factors:(1) Good cluster quality for successfully generating the demanded sets of images (2) Good image quality for high-quality image generation (3) The ability to access and generate the data.
To evaluate the quality of unsupervised conditional image generation, we measure performance on two axes: cluster quality and image quality. We estimate clusters by optimizing \eqref{eqn:cluster_mcr}, and show results and comparisons with both recent and classical methods in Table  \ref{tab:comparsion_ucig}. Training details of our method for the additional MLP head can be found in the Appendix \ref{sec:appendix_networkarch}. 

\vspace{-1mm}
\textbf{Cluster Quality.} We measure normalized mutual information (NMI) and clustering accuracy for cluster quality on CIFAR-10 clustered into 10 classes and CIFAR-100(20), which is clustered into 20 super-classes. From Table  \ref{tab:comparsion_ucig}, we observe that on CIFAR-10, \ours{} results in an NMI that is almost double that of the existing SOTA on both GAN-based and VAE-based methods, with significantly improved clustering accuracy.
Unlike many baselines, we also demonstrate that our method scales to the more challenging CIFAR-100(20) dataset, where it also significantly outperforms alternatives.
%On more difficult datasets like CIFAR-100(20), as many of the existing methods struggle to scale up to more complicated dataset.
%We have achieved much better NMI and Accuracy comparing to the ones that do scale up.
Our improved clustering quality suggests potential for improving unsupervised conditional image generation, which relies on first finding statistically (and hence visually) meaningful clusters.
% This has tremendously improved generative model's ability to perform unsupervised conditional image generation, because it heavily relies on finding the correct (statistcally hence visually meaningful) clusters first!

\vspace{-1mm}
\textbf{Image Quality.} 
We use Frechet Inception Distance (FID) \citep{heusel2017gans} and Inception Score (IS) \citep{salimans2016improved} to measure image quality. From Table  \ref{tab:comparsion_ucig}, it is evident that \ours{} maintains competitive image quality compared to other methods, measured both by FID and IS. We also compare original images against reconstructed ones in Figure \ref{fig:vis_recon}, where we see that the original $\X$ is very similar to the reconstructed $\hat{\X}$; \ours{} indeed achieves very good sample-wise self-consistency.

\begin{figure}[t]
    \footnotesize
    \centering
    \includegraphics[width=0.915\textwidth]{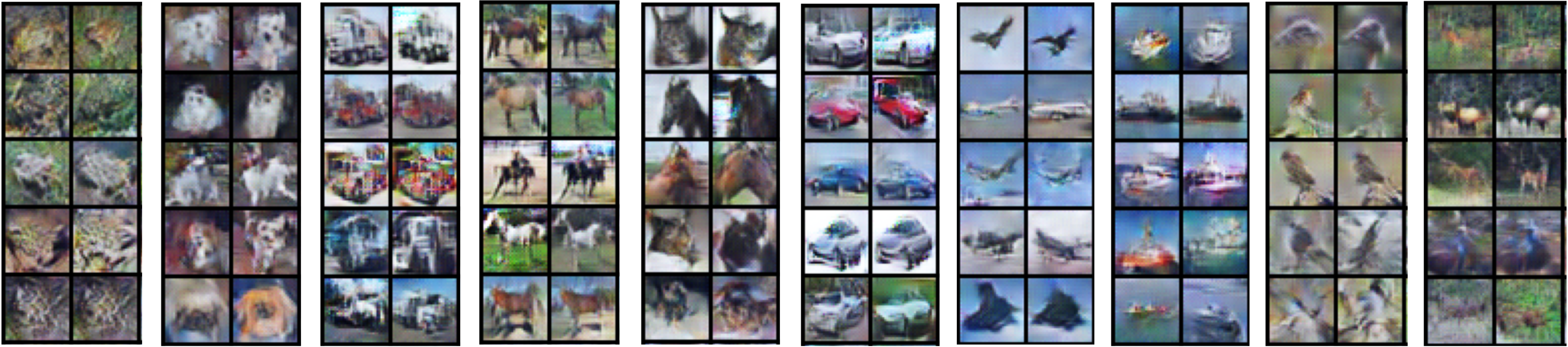}
    \caption{\small Unsupervised conditional image generation from each cluster of CIFAR-10, using \ours{}. Images from different rows mean generation from different principal components of each cluster.\vspace{-2mm}}
    \label{fig:vis_clustering}
\end{figure}

\vspace{-1mm}
\textbf{Unsupervised Conditional Image Generation.}
In Figure  \ref{fig:vis_clustering}, we  visualize images generated from the ten unsupervised clusters from \eqref{eqn:cluster_mcr}. Each block represents one cluster and each row represents one principal component for each cluster. Despite learning and training without labels, the model not only organizes samples into correct clusters, but is also able to preserve statistical diversities within each cluster/class. We can easily recover the diversity within each cluster by computing different principal components and then sample and generate accordingly! More detailed illustrations with more samples is provided in Appendix \ref{sec:appendix_ucig}.       

\begin{table}[t]
    \centering
    \small
    \setlength{\tabcolsep}{3.5pt}
    \renewcommand{\arraystretch}{1.25}
    \begin{tabular}{l|cccc|cccc}
    \multirow{2}{*}{Method} & \multicolumn{4}{c|}{CIFAR-10} & \multicolumn{4}{c}{CIFAR-100(20)} \\
    \cline{2-9}
     ~  & NMI & Accuracy & FID$\downarrow$ & IS$\uparrow$ & NMI & Accuracy & FID$\downarrow$ & IS$\uparrow$ \\
    \hline
    \hline
    \textit{GAN based methods}          &  ~   & ~    & ~    & ~    & ~    & ~ & ~    & ~  \\
    Self-Conditioned GAN \citep{liu2020diverse}&  0.333    & 0.117  & 18.0    & 7.7    & 0.214   & 0.092  & 24.1   & 5.2   \\
    SLOGAN \citep{hwang2021stein}&  0.340    & -  & 20.6    & -    & -   &  -  & -   &  -  \\
    
    \hline
    \textit{VAE based methods}    & ~       & ~    & ~    & ~    & ~    & ~  & ~    & ~  \\
    GMVAE\citep{dilokthanakul2016deep} & - & 0.247 & -  & -   & -   & - & -   &  - \\
    Variational Clustering         & -     & 0.445 & -   & -   & -   & - & -   &  - \\

    \hline
    \textit{CTRL based methods}    & ~       & ~    & ~    & ~    & ~    & ~  & ~    & ~  \\
    \ours{} (ours)                  & \textbf{0.658}     & \textbf{0.799} & \textbf{17.4}   & \textbf{8.1}   & \textbf{0.374}   & \textbf{0.433} & \textbf{20.1}   &  \textbf{7.7} \\
    \end{tabular}      
    % \vspace{-0.1in}
    \caption{
        \small Comparison of the quality of UCIG on CIFAR-10 and CIFAR-100(20).
        Many of the methods compared do not provide code that scales up to CIFAR-100(20), in which case we leave the corresponding table cell blank.\vspace{-6mm} }
    \label{tab:comparsion_ucig}
\end{table}

\vspace{-2mm}
\subsection{Benefits of \ours{}'s Structured Representation}\label{sec:exp_benefit}
\vspace{-2mm}
As shown in the previous section, on datasets like CIFAR-10, CIFAR-100, and Tiny-ImageNet, our framework is able to achieve representation quality on par with the best discriminative self-supervised learning methods.
A clear advantage of this is computational efficiency; only a single representation needs to be trained for a much broader set of tasks.
This subsection aims to provide additional insights on how a unified model can be more beneficial for a broader range of tasks.

\vspace{-1mm}
\textbf{Domain Transfer.}
Regenerating images is demanding on the encoder, which is required to produce a more informative representation than contrastive training would. We hypothesize that the encoder trained with generative task may retain more information about the image and allow the representation to generalize better. To verify this, we compare the accuracy on CIFAR-100 using models learned from  CIFAR-10 in Table  \ref{tab:comparison_transfer}. When compared to purely discriminative self-supervised learning models, we observe that \ours{} is ~4 percent better than other methods on classification accuracy.

\begin{figure}[h!]
     \footnotesize
     \centering
     \subfigure[\ours{}]{
         \includegraphics[width=0.455\textwidth]{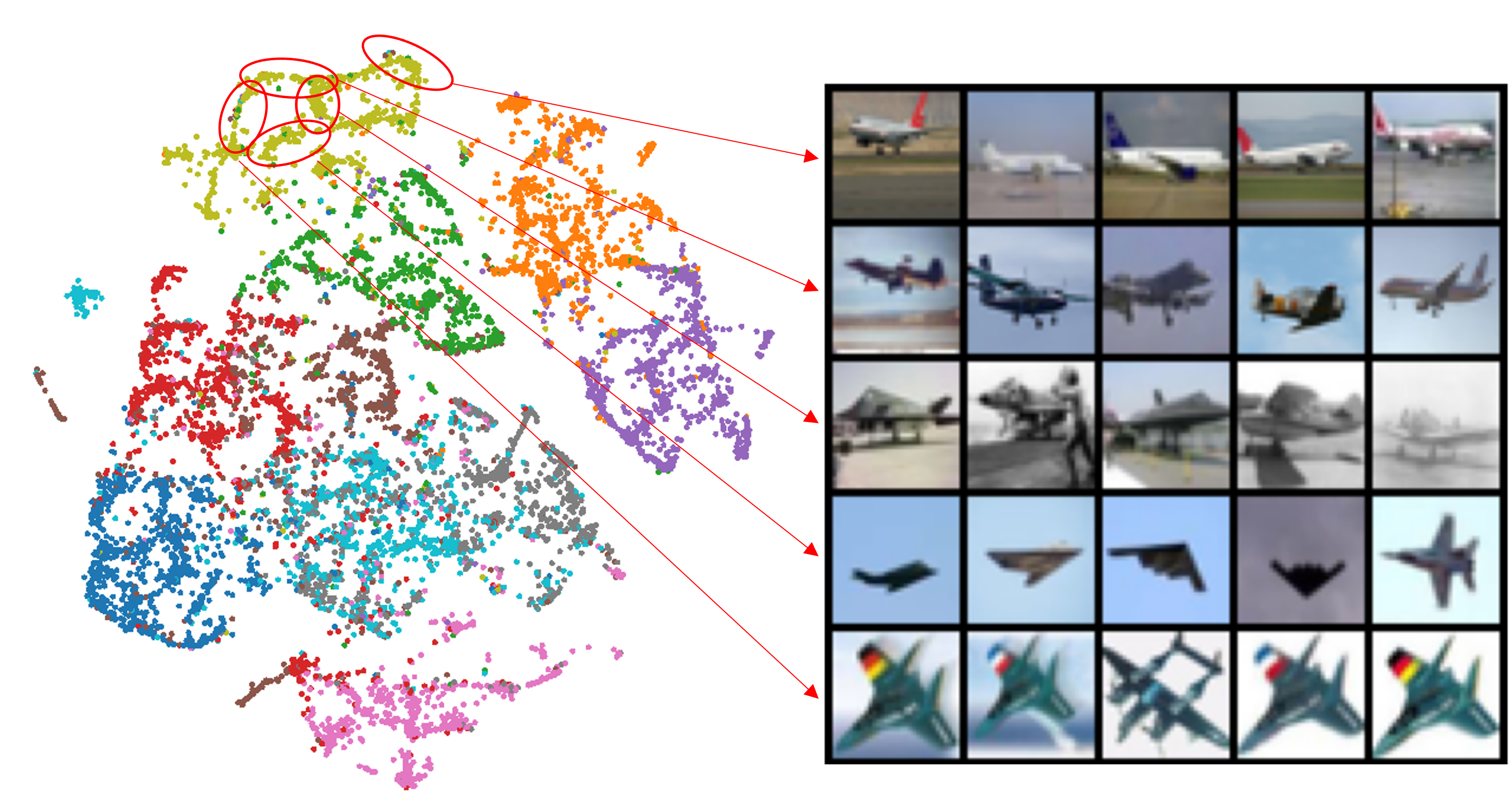}
     }
     \subfigure[MoCoV2]{
         \includegraphics[width=0.455\textwidth]{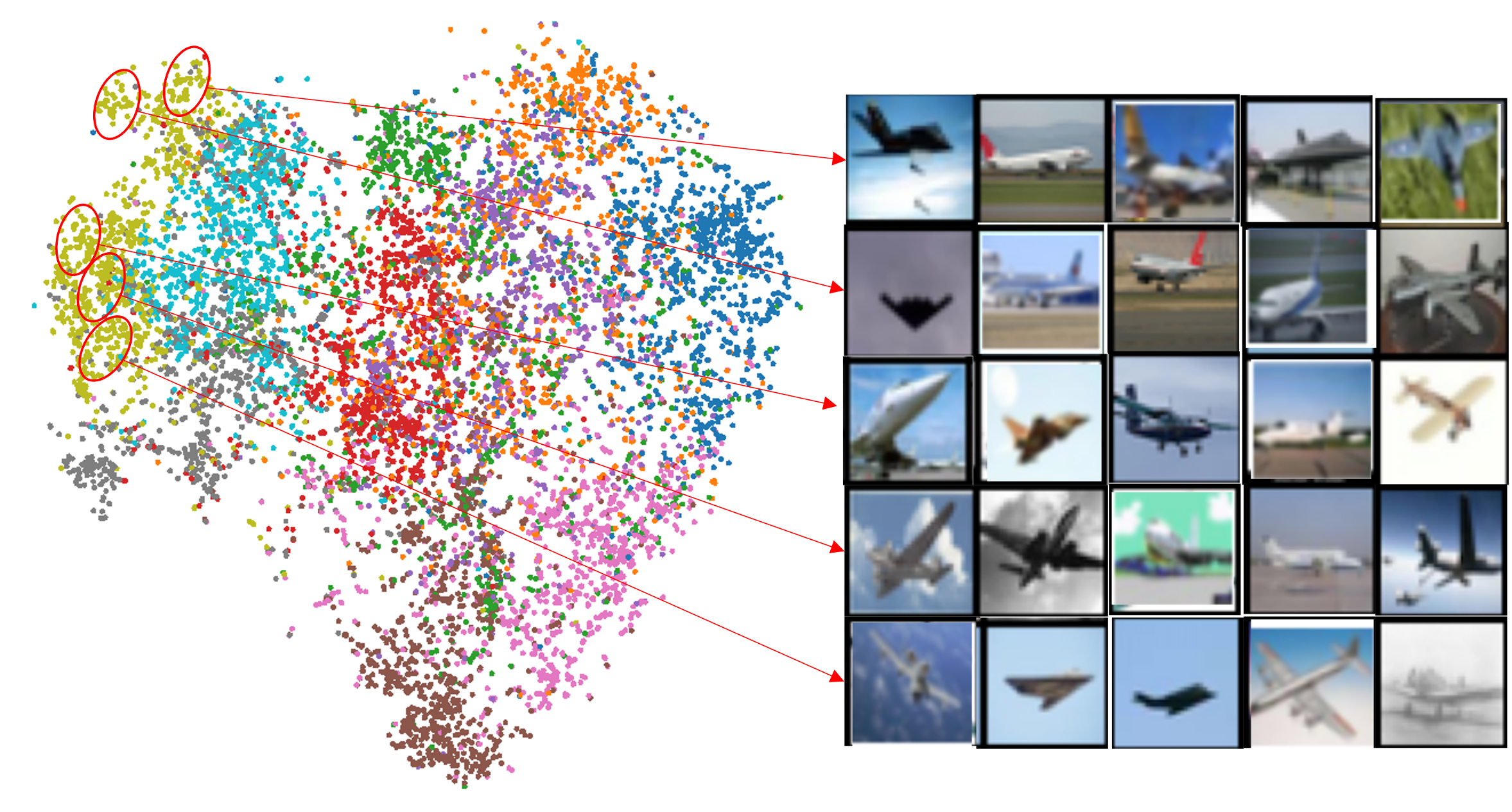}
     }
    \vspace{-0.15in}
    \caption{\small t-SNE visualizations of learned features of CIFAR-10 with different models.} \vspace{-5mm}
    \label{fig:tsne}
\end{figure}

\vspace{-1mm}
\textbf{Visualization of Emerged Structures.} The representations learned by \ours{} are significantly different from those learned from previous either discriminative and generative methods. To illustrate this, we use t-SNE  \citep{van2008visualizing} to visualize the learned representation in 2D. Figure \ref{fig:tsne} compares the t-SNE of representations learned for CIFAR-10 by \ours{} and MoCoV2, respectively. It is clear that the representation learned by \ours{} are much more structured and better organized: classes are more evident, and  features within each class form clear piecewise linear structures.
\vspace{-5mm}
\begin{table}[h!]
\begin{small}
    \centering
    \setlength{\tabcolsep}{4.5pt}
    \renewcommand{\arraystretch}{1.25}
    \begin{tabular}{l|c|c|c|c}
    Method  & SIMCLR   & MoCoV2  & BYOL & \ours{}\\
    \hline
    \hline
    Accuracy    & 0.422 & 0.436  & 0.437 & 0.481  \\  
    \end{tabular}      
    % \vspace{-0.1in}
    \caption{\small Comparing the transfer ability with purely discriminative self-supervised learning methods. All methods are trained unsupervised on CIFAR-10 and tested on CIFAR-100. \vspace{-7mm}}
     \label{tab:comparison_transfer}
\end{small}     
\end{table}
% \vspace{-5mm}

\vspace{-2mm}
\section{Conclusion and Discussion} 
\vspace{-4mm}
In this work, we proposed an unsupervised formulation of the closed-loop transcription framework~\citep{dai2022ctrl}. We experimentally demonstrate that it is possible to learn a unified representation for both discriminative and generative purposes, resulting in highly structured representations. Further, we show that these two purposes mutually benefit each other in various tasks, e.g.,  conditional image generation and domain tranfers. Compared to the more specialized representations learned in prior works, our results suggest that such a unified representation has the potential in supporting and benefiting a wider range of new tasks. In future work, we believe the learned representations can be further improved by jointly optimizing the feature representation and feature clusters, as suggested in the original rate reduction paper \citep{chan2021redunet}. Features with high likelihood of belonging to the same cluster can be further linearized and compressed. Due to its unifying nature and the simplicity of the underlying concepts, this new framework may be extended beyond image data, such as sequential or dynamical observations.

\section{Acknowledgements}

Yi Ma acknowledges support from ONR grants N00014-20-1-2002 and N00014-22-1-2102, the joint Simons Foundation-NSF DMS grant \#2031899, as well as partial support from Berkeley FHL Vive Center for Enhanced Reality and Berkeley Center for Augmented Cognition, Tsinghua-Berkeley Shenzhen Institute (TBSI) Research Fund, and Berkeley AI Research (BAIR).

% \newpage
% \section*{Ethics Statement}
% All authors agree and will adhere to the conference's Code of Ethics. We do not anticipate any potential ethics issues regarding the research conducted in this work. 

% \section*{Reproducibility Statement}
% Settings and implementation details of network architectures, optimization methods, and some common hyper-parameters are described in the Appendix~\ref{sec:appendix_networkarch}. We will also make our source code available upon request by the reviewers or the area chairs.

\bibliographystyle{iclr2023_conference}

\begin{thebibliography}{48}
\providecommand{\natexlab}[1]{#1}
\providecommand{\url}[1]{\texttt{#1}}
\expandafter\ifx\csname urlstyle\endcsname\relax
  \providecommand{\doi}[1]{doi: #1}\else
  \providecommand{\doi}{doi: \begingroup \urlstyle{rm}\Url}\fi

\bibitem[Antoniou et~al.(2017)Antoniou, Storkey, and Edwards]{antoniou2017data}
Antreas Antoniou, Amos Storkey, and Harrison Edwards.
\newblock Data augmentation generative adversarial networks.
\newblock \emph{arXiv preprint arXiv:1711.04340}, 2017.

\bibitem[Bardes et~al.(2021)Bardes, Ponce, and LeCun]{bardes2021vicreg}
Adrien Bardes, Jean Ponce, and Yann LeCun.
\newblock Vicreg: Variance-invariance-covariance regularization for
  self-supervised learning.
\newblock \emph{arXiv preprint arXiv:2105.04906}, 2021.

\bibitem[Chan et~al.(2022)Chan, Yu, You, Qi, Wright, and Ma]{chan2021redunet}
Kwan Ho~Ryan Chan, Yaodong Yu, Chong You, Haozhi Qi, John Wright, and Yi~Ma.
\newblock Redu{N}et: A white-box deep network from the principle of maximizing
  rate reduction.
\newblock \emph{Journal of Machine Learning Research}, 23\penalty0
  (114):\penalty0 1--103, 2022.
\newblock URL \url{http://jmlr.org/papers/v23/21-0631.html}.

\bibitem[Chen et~al.(2020{\natexlab{a}})Chen, Radford, Child, Wu, Jun, Luan,
  and Sutskever]{chen2020generative}
Mark Chen, Alec Radford, Rewon Child, Jeffrey Wu, Heewoo Jun, David Luan, and
  Ilya Sutskever.
\newblock Generative pretraining from pixels.
\newblock In \emph{International Conference on Machine Learning}, pp.\
  1691--1703. PMLR, 2020{\natexlab{a}}.

\bibitem[Chen et~al.(2019)Chen, Zhai, Ritter, Lucic, and Houlsby]{chen2019self}
Ting Chen, Xiaohua Zhai, Marvin Ritter, Mario Lucic, and Neil Houlsby.
\newblock Self-supervised gans via auxiliary rotation loss.
\newblock In \emph{Proceedings of the IEEE/CVF conference on computer vision
  and pattern recognition}, pp.\  12154--12163, 2019.

\bibitem[Chen et~al.(2020{\natexlab{b}})Chen, Kornblith, Norouzi, and
  Hinton]{chen2020simple}
Ting Chen, Simon Kornblith, Mohammad Norouzi, and Geoffrey Hinton.
\newblock A simple framework for contrastive learning of visual
  representations.
\newblock In \emph{International conference on machine learning}, pp.\
  1597--1607. PMLR, 2020{\natexlab{b}}.

\bibitem[Chen et~al.(2016)Chen, Duan, Houthooft, Schulman, Sutskever, and
  Abbeel]{chen2016infogan}
Xi~Chen, Yan Duan, Rein Houthooft, John Schulman, Ilya Sutskever, and Pieter
  Abbeel.
\newblock Infogan: Interpretable representation learning by information
  maximizing generative adversarial nets.
\newblock \emph{Advances in neural information processing systems}, 29, 2016.

\bibitem[Chen et~al.(2022)Chen, Ding, Wang, Xin, Mo, Wang, Han, Luo, Zeng, and
  Wang]{chen2022context}
Xiaokang Chen, Mingyu Ding, Xiaodi Wang, Ying Xin, Shentong Mo, Yunhao Wang,
  Shumin Han, Ping Luo, Gang Zeng, and Jingdong Wang.
\newblock Context autoencoder for self-supervised representation learning.
\newblock \emph{arXiv preprint arXiv:2202.03026}, 2022.

\bibitem[Chen \& He(2020)Chen and He]{chen2020simsiam}
Xinlei Chen and Kaiming He.
\newblock Exploring simple siamese representation learning.
\newblock In \emph{CVPR}, 2020.

\bibitem[Dai et~al.(2022)Dai, Tong, Li, Wu, Psenka, Chan, Zhai, Yu, Yuan, Shum,
  et~al.]{dai2022ctrl}
Xili Dai, Shengbang Tong, Mingyang Li, Ziyang Wu, Michael Psenka, Kwan Ho~Ryan
  Chan, Pengyuan Zhai, Yaodong Yu, Xiaojun Yuan, Heung-Yeung Shum, et~al.
\newblock Ctrl: Closed-loop transcription to an ldr via minimaxing rate
  reduction.
\newblock \emph{Entropy}, 24\penalty0 (4):\penalty0 456, 2022.

\bibitem[Deng et~al.(2009)Deng, Dong, Socher, Li, Li, and
  Fei-Fei]{deng2009imagenet}
Jia Deng, Wei Dong, Richard Socher, Li-Jia Li, Kai Li, and Li~Fei-Fei.
\newblock Imagenet: A large-scale hierarchical image database.
\newblock In \emph{2009 IEEE conference on computer vision and pattern
  recognition}, pp.\  248--255. Ieee, 2009.

\bibitem[Dilokthanakul et~al.(2016)Dilokthanakul, Mediano, Garnelo, Lee,
  Salimbeni, Arulkumaran, and Shanahan]{dilokthanakul2016deep}
Nat Dilokthanakul, Pedro~AM Mediano, Marta Garnelo, Matthew~CH Lee, Hugh
  Salimbeni, Kai Arulkumaran, and Murray Shanahan.
\newblock Deep unsupervised clustering with gaussian mixture variational
  autoencoders.
\newblock \emph{arXiv preprint arXiv:1611.02648}, 2016.

\bibitem[Donahue et~al.(2016)Donahue, Kr{\"a}henb{\"u}hl, and
  Darrell]{donahue2016adversarial}
Jeff Donahue, Philipp Kr{\"a}henb{\"u}hl, and Trevor Darrell.
\newblock Adversarial feature learning.
\newblock \emph{arXiv preprint arXiv:1605.09782}, 2016.

\bibitem[Dumoulin et~al.(2016)Dumoulin, Belghazi, Poole, Mastropietro, Lamb,
  Arjovsky, and Courville]{dumoulin2016adversarially}
Vincent Dumoulin, Ishmael Belghazi, Ben Poole, Olivier Mastropietro, Alex Lamb,
  Martin Arjovsky, and Aaron Courville.
\newblock Adversarially learned inference.
\newblock \emph{arXiv preprint arXiv:1606.00704}, 2016.

\bibitem[Falcon et~al.(2021)Falcon, Jha, Koker, and Cho]{falcon2021aavae}
William Falcon, Ananya~Harsh Jha, Teddy Koker, and Kyunghyun Cho.
\newblock Aavae: Augmentation-augmented variational autoencoders.
\newblock \emph{arXiv preprint arXiv:2107.12329}, 2021.

\bibitem[Goodfellow et~al.(2014)Goodfellow, Pouget-Abadie, Mirza, Xu,
  Warde-Farley, Ozair, Courville, and Bengio]{goodfellow2014generative}
Ian Goodfellow, Jean Pouget-Abadie, Mehdi Mirza, Bing Xu, David Warde-Farley,
  Sherjil Ozair, Aaron Courville, and Yoshua Bengio.
\newblock Generative adversarial nets.
\newblock \emph{Advances in neural information processing systems}, 27, 2014.

\bibitem[Grill et~al.(2020{\natexlab{a}})Grill, Strub, Altch{\'e}, Tallec,
  Richemond, Buchatskaya, Doersch, Avila~Pires, Guo, Gheshlaghi~Azar,
  et~al.]{grill2020bootstrap}
Jean-Bastien Grill, Florian Strub, Florent Altch{\'e}, Corentin Tallec, Pierre
  Richemond, Elena Buchatskaya, Carl Doersch, Bernardo Avila~Pires, Zhaohan
  Guo, Mohammad Gheshlaghi~Azar, et~al.
\newblock Bootstrap your own latent-a new approach to self-supervised learning.
\newblock \emph{Advances in Neural Information Processing Systems},
  33:\penalty0 21271--21284, 2020{\natexlab{a}}.

\bibitem[Grill et~al.(2020{\natexlab{b}})Grill, Strub, Altché, Tallec,
  Richemond, Buchatskaya, Doersch, Pires, Guo, Azar, Piot, Kavukcuoglu, Munos,
  and Valko]{grill2020byol}
Jean-Bastien Grill, Florian Strub, Florent Altché, Corentin Tallec, Pierre~H.
  Richemond, Elena Buchatskaya, Carl Doersch, Bernardo~Avila Pires,
  Zhaohan~Daniel Guo, Mohammad~Gheshlaghi Azar, Bilal Piot, Koray Kavukcuoglu,
  Rémi Munos, and Michal Valko.
\newblock Bootstrap your own latent: A new approach to self-supervised
  learning.
\newblock In \emph{NeurIPS}, 2020{\natexlab{b}}.

\bibitem[Gupta et~al.(2020)Gupta, Singh, and Shrivastava]{gupta2020patchvae}
Kamal Gupta, Saurabh Singh, and Abhinav Shrivastava.
\newblock Patchvae: Learning local latent codes for recognition.
\newblock In \emph{Proceedings of the IEEE/CVF Conference on Computer Vision
  and Pattern Recognition}, pp.\  4746--4755, 2020.

\bibitem[He et~al.(2020)He, Fan, Wu, Xie, and Girshick]{he2020momentum}
Kaiming He, Haoqi Fan, Yuxin Wu, Saining Xie, and Ross Girshick.
\newblock Momentum contrast for unsupervised visual representation learning.
\newblock In \emph{Proceedings of the IEEE/CVF conference on computer vision
  and pattern recognition}, pp.\  9729--9738, 2020.

\bibitem[He et~al.(2021)He, Chen, Xie, Li, Doll{\'a}r, and
  Girshick]{he2021masked}
Kaiming He, Xinlei Chen, Saining Xie, Yanghao Li, Piotr Doll{\'a}r, and Ross
  Girshick.
\newblock Masked autoencoders are scalable vision learners.
\newblock \emph{arXiv preprint arXiv:2111.06377}, 2021.

\bibitem[Heusel et~al.(2017)Heusel, Ramsauer, Unterthiner, Nessler, and
  Hochreiter]{heusel2017gans}
Martin Heusel, Hubert Ramsauer, Thomas Unterthiner, Bernhard Nessler, and Sepp
  Hochreiter.
\newblock Gans trained by a two time-scale update rule converge to a local nash
  equilibrium.
\newblock \emph{Advances in neural information processing systems}, 30, 2017.

\bibitem[Higgins et~al.(2016)Higgins, Matthey, Pal, Burgess, Glorot, Botvinick,
  Mohamed, and Lerchner]{higgins2016beta}
Irina Higgins, Loic Matthey, Arka Pal, Christopher Burgess, Xavier Glorot,
  Matthew Botvinick, Shakir Mohamed, and Alexander Lerchner.
\newblock beta-vae: Learning basic visual concepts with a constrained
  variational framework.
\newblock \emph{arXiv preprint arXiv:1804.03599}, 2016.

\bibitem[Hou et~al.(2021)Hou, Shen, Cao, and Cheng]{hou2021self}
Liang Hou, Huawei Shen, Qi~Cao, and Xueqi Cheng.
\newblock Self-supervised gans with label augmentation.
\newblock \emph{Advances in Neural Information Processing Systems}, 34, 2021.

\bibitem[Hwang et~al.(2021)Hwang, Kim, Jung, Jang, Lee, and
  Yoon]{hwang2021stein}
Uiwon Hwang, Heeseung Kim, Dahuin Jung, Hyemi Jang, Hyungyu Lee, and Sungroh
  Yoon.
\newblock Stein latent optimization for generative adversarial networks.
\newblock \emph{arXiv preprint arXiv:2106.05319}, 2021.

\bibitem[Jeong \& Shin(2021)Jeong and Shin]{jeong2021training}
Jongheon Jeong and Jinwoo Shin.
\newblock Training gans with stronger augmentations via contrastive
  discriminator.
\newblock \emph{arXiv preprint arXiv:2103.09742}, 2021.

\bibitem[Jiang et~al.(2016)Jiang, Zheng, Tan, Tang, and
  Zhou]{jiang2016variational}
Zhuxi Jiang, Yin Zheng, Huachun Tan, Bangsheng Tang, and Hanning Zhou.
\newblock Variational deep embedding: An unsupervised and generative approach
  to clustering.
\newblock \emph{arXiv preprint arXiv:1611.05148}, 2016.

\bibitem[Josselyn \& Tonegawa(2020)Josselyn and Tonegawa]{Josselyn2020MemoryER}
Sheena~A. Josselyn and Susumu Tonegawa.
\newblock Memory engrams: Recalling the past and imagining the future.
\newblock \emph{Science}, 367, 2020.

\bibitem[Keller \& Mrsic-Flogel(2018)Keller and Mrsic-Flogel]{Keller2018-ez}
Georg~B Keller and Thomas~D Mrsic-Flogel.
\newblock Predictive processing: A canonical cortical computation.
\newblock \emph{Neuron}, 100\penalty0 (2):\penalty0 424--435, October 2018.

\bibitem[Kim et~al.(2021)Kim, Kim, and Lee]{kim2021hybridiGPT}
Saehoon Kim, Sungwoong Kim, and Juho Lee.
\newblock Hybrid generative-contrastive representation learning.
\newblock \emph{arXiv preprint arXiv:2106.06162}, 2021.

\bibitem[Kingma \& Ba(2014)Kingma and Ba]{kingma2014adam}
Diederik~P Kingma and Jimmy Ba.
\newblock Adam: A method for stochastic optimization.
\newblock \emph{arXiv preprint arXiv:1412.6980}, 2014.

\bibitem[Kingma \& Welling(2013)Kingma and Welling]{kingma2013auto}
Diederik~P Kingma and Max Welling.
\newblock Auto-encoding variational bayes.
\newblock \emph{arXiv preprint arXiv:1312.6114}, 2013.

\bibitem[Krizhevsky et~al.(2014)Krizhevsky, Nair, and
  Hinton]{krizhevsky2014cifar}
Alex Krizhevsky, Vinod Nair, and Geoffrey Hinton.
\newblock The {CIFAR}-10 dataset.
\newblock \emph{online: http://www.cs.toronto.edu/kriz/cifar.html}, 55, 2014.

\bibitem[Krizhevsky et~al.(2009)]{krizhevsky2009learning}
Alex Krizhevsky et~al.
\newblock Learning multiple layers of features from tiny images.
\newblock \emph{arXiv preprint arXiv:1312.6114}, 2009.

\bibitem[Li et~al.(2022)Li, Chen, LeCun, and Sommer]{li2022neural}
Zengyi Li, Yubei Chen, Yann LeCun, and Friedrich~T Sommer.
\newblock Neural manifold clustering and embedding.
\newblock \emph{arXiv preprint arXiv:2201.10000}, 2022.

\bibitem[Liu et~al.(2020)Liu, Wang, Bau, Zhu, and Torralba]{liu2020diverse}
Steven Liu, Tongzhou Wang, David Bau, Jun-Yan Zhu, and Antonio Torralba.
\newblock Diverse image generation via self-conditioned gans.
\newblock In \emph{Proceedings of the IEEE/CVF conference on computer vision
  and pattern recognition}, pp.\  14286--14295, 2020.

\bibitem[Ma et~al.(2007)Ma, Derksen, Hong, and Wright]{ma2007segmentation}
Yi~Ma, Harm Derksen, Wei Hong, and John Wright.
\newblock Segmentation of multivariate mixed data via lossy data coding and
  compression.
\newblock \emph{PAMI}, 2007.

\bibitem[Mukherjee et~al.(2019)Mukherjee, Asnani, Lin, and
  Kannan]{mukherjee2019clustergan}
Sudipto Mukherjee, Himanshu Asnani, Eugene Lin, and Sreeram Kannan.
\newblock Clustergan: Latent space clustering in generative adversarial
  networks.
\newblock In \emph{Proceedings of the AAAI conference on artificial
  intelligence}, volume~33, pp.\  4610--4617, 2019.

\bibitem[Parmar et~al.(2021)Parmar, Li, Lee, and Tu]{parmar2021dual}
Gaurav Parmar, Dacheng Li, Kwonjoon Lee, and Zhuowen Tu.
\newblock Dual contradistinctive generative autoencoder.
\newblock In \emph{Proceedings of the IEEE/CVF Conference on Computer Vision
  and Pattern Recognition}, pp.\  823--832, 2021.

\bibitem[Prasad et~al.(2020)Prasad, Das, and Bhowmick]{prasad2020variational}
Vignesh Prasad, Dipanjan Das, and Brojeshwar Bhowmick.
\newblock Variational clustering: Leveraging variational autoencoders for image
  clustering.
\newblock In \emph{2020 International Joint Conference on Neural Networks
  (IJCNN)}, pp.\  1--10. IEEE, 2020.

\bibitem[Radford et~al.(2015)Radford, Metz, and
  Chintala]{radford2015unsupervised}
Alec Radford, Luke Metz, and Soumith Chintala.
\newblock Unsupervised representation learning with deep convolutional
  generative adversarial networks.
\newblock \emph{arXiv preprint arXiv:1511.06434}, 2015.

\bibitem[Salimans et~al.(2016)Salimans, Goodfellow, Zaremba, Cheung, Radford,
  and Chen]{salimans2016improved}
Tim Salimans, Ian Goodfellow, Wojciech Zaremba, Vicki Cheung, Alec Radford, and
  Xi~Chen.
\newblock Improved techniques for training gans.
\newblock \emph{Advances in neural information processing systems}, 29, 2016.

\bibitem[Tong et~al.(2022)Tong, Dai, Wu, Li, Yi, and Ma]{tong2022incremental}
Shengbang Tong, Xili Dai, Ziyang Wu, Mingyang Li, Brent Yi, and Yi~Ma.
\newblock Incremental learning of structured memory via closed-loop
  transcription.
\newblock \emph{arXiv:2202.05411}, 2022.

\bibitem[Van~der Maaten \& Hinton(2008)Van~der Maaten and
  Hinton]{van2008visualizing}
Laurens Van~der Maaten and Geoffrey Hinton.
\newblock Visualizing data using t-sne.
\newblock \emph{Journal of machine learning research}, 9\penalty0 (11), 2008.

\bibitem[Ven et~al.(2020)Ven, Siegelmann, Tolias, et~al.]{2020Vandeven}
Gido~M Ven, Hava~T Siegelmann, Andreas~S Tolias, et~al.
\newblock Brain-inspired replay for continual learning with artificial neural
  networks.
\newblock \emph{Nature Communications}, 11\penalty0 (1):\penalty0 1--14, 2020.

\bibitem[Wang et~al.(2004)Wang, Bovik, Sheikh, and
  Simoncelli]{image-similarity}
Zhou Wang, A.C. Bovik, H.R. Sheikh, and E.P. Simoncelli.
\newblock Image quality assessment: from error visibility to structural
  similarity.
\newblock \emph{IEEE Transactions on Image Processing}, 13\penalty0
  (4):\penalty0 600--612, 2004.
\newblock \doi{10.1109/TIP.2003.819861}.

\bibitem[Yu et~al.(2020)Yu, Chan, You, Song, and Ma]{yu2020learning}
Yaodong Yu, Kwan Ho~Ryan Chan, Chong You, Chaobing Song, and Yi~Ma.
\newblock Learning diverse and discriminative representations via the principle
  of maximal coding rate reduction.
\newblock \emph{Advances in Neural Information Processing Systems},
  33:\penalty0 9422--9434, 2020.

\bibitem[Zbontar et~al.(2021)Zbontar, Jing, Misra, LeCun, and
  Deny]{zbontar2021barlow}
Jure Zbontar, Li~Jing, Ishan Misra, Yann LeCun, and St{\'e}phane Deny.
\newblock Barlow twins: Self-supervised learning via redundancy reduction.
\newblock In \emph{International Conference on Machine Learning}, pp.\
  12310--12320. PMLR, 2021.

\end{thebibliography}
\newpage

\appendix

\section{Training Details}\label{sec:appendix_networkarch}
\subsection{Network Architectures}

\begin{figure}[ht]
     \footnotesize
     \centering
     \subfigure[ResBlock Up architecture]{
         \includegraphics[width=0.305\textwidth]{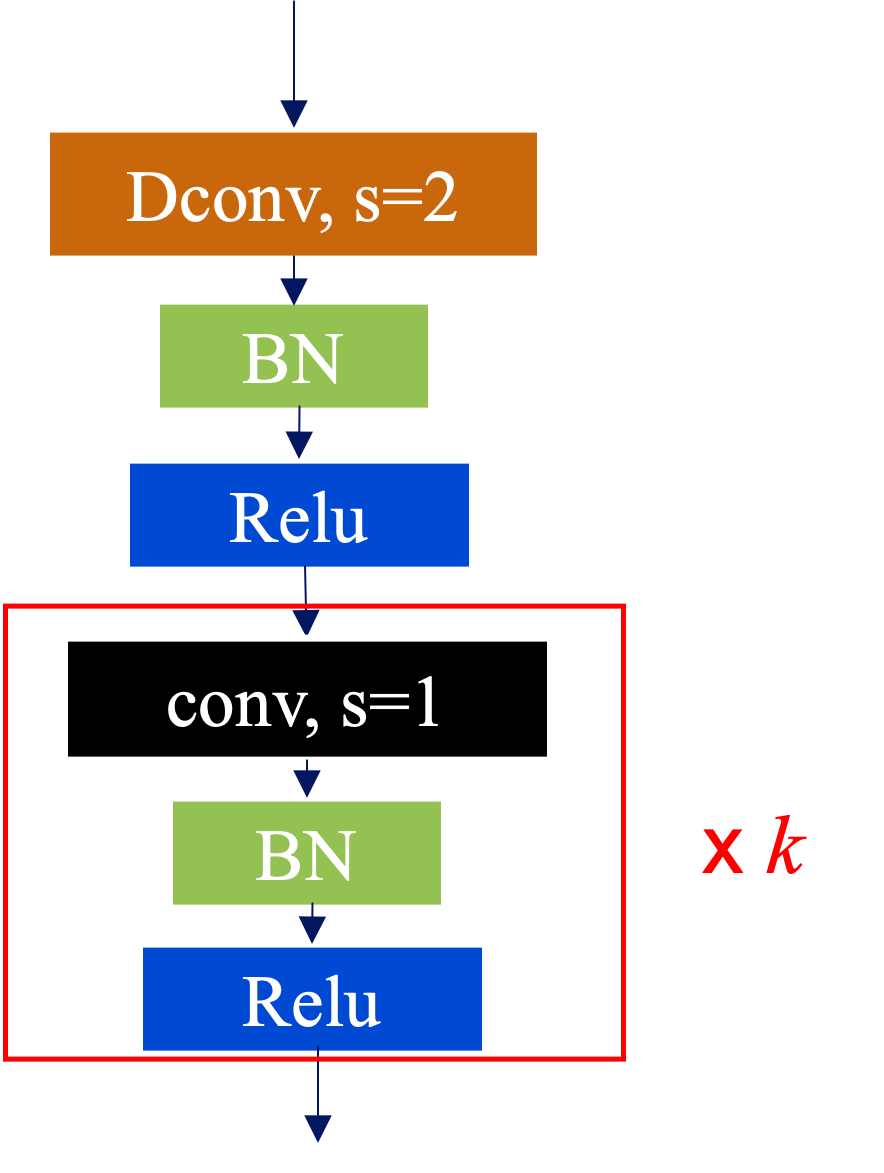}
     }
     \subfigure[ResBlock Down architecture]{
         \includegraphics[width=0.315\textwidth]{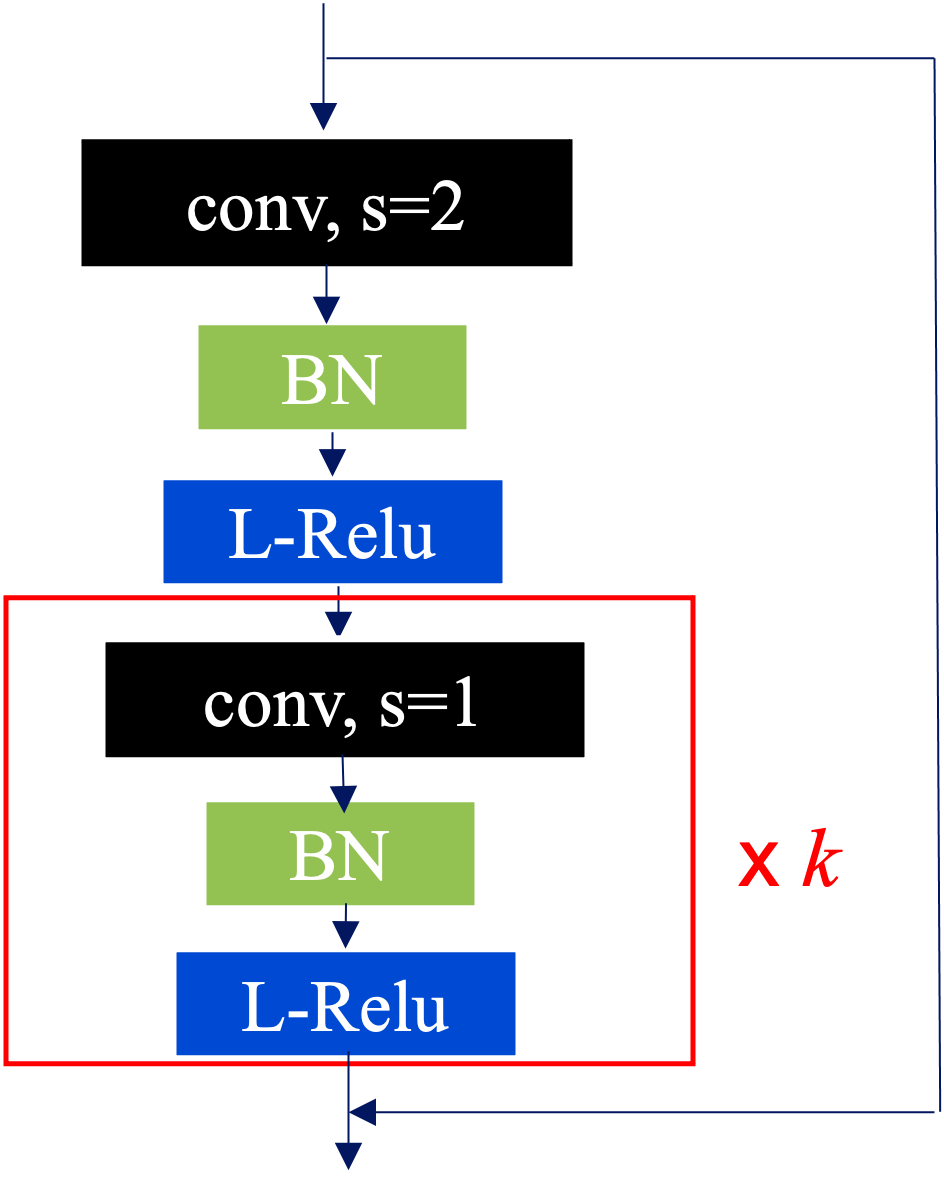}
     }
    \vspace{-0.15in}
    \caption{\small Architecture of two ResBlock.  \vspace{-6mm}}
    \label{fig:resblock}
\end{figure}

Table~\ref{arch:decoder}, \ref{arch:encoder} and Figure~\ref{fig:resblock} give details on the network architecture for the decoder and the encoder networks used for experiments. The black rectangle marked with "conv, s=2" means a convlutional layer with stride 2. The orange rectangle marked with "dconv, s=2" means a deconvolutional layer with stride 2. The "x k" besides red frame means we regard these layers in red frame as a block and stack it k times. All $\alpha$ values in Leaky-ReLU (i.e. lReLU) of the encoder are set to $0.2$. We set ($nz=128$, $nc=3$, $k=3$) for CIFAR-10, ($nz=256$, $nc=3$, $k=4$) for CIFAR-100, and ($nz=256$, $nc=3$, $k=4$) for Tiny-ImageNet. As a comparison, ResNet-18 contains around 11 million parameters, whereas our encoder only contains between 4 and 6 million parameters depending on the choice of k. 

Table~\ref{arch:linear_classifier} gives details of the network architecture for the linear classifier and Table \ref{arch:mlp_ucig} gives details of the network architecture for the additional MLP head used for unsupervised conditional image generation training.

\begin{table}[ht]
 \setlength{\tabcolsep}{0.5cm} 
 \centering
 \begin{tabular}{c}
 \hline
 \hline
 $\z \in \R^{1 \times 1 \times nz}$  \\
 \hline
 ResBlockUp. 256 \\
 \hline
 ResBlockUp. 128 \\
 \hline
 ResBlockUp. 64  \\
 \hline
 4 $\times$ 4, stride=2, pad=1 deconv. 1 Tanh   \\
 \hline
 \hline
 \end{tabular}
 \caption{Network architecture of the decoder $g(\cdot, \eta)$.}
 \label{arch:decoder}
\end{table}

\begin{table}[ht]
\setlength{\tabcolsep}{0.5cm}
\centering
 \begin{tabular}{c}
 \hline
 \hline
 Image $\x \in \R^{32 \times 32 \times nc}$  \\
 \hline
 ResBlockDown 64    \\
 \hline
 ResBlockDown 128   \\
 \hline
 ResBlockDown 256   \\
 \hline
 4 $\times$ 4, stride=1, pad=0 conv $nz$   \\
 \hline
 \hline
 \end{tabular}
\caption{Network architecture of the encoder $f(\cdot, \theta)$.}
\label{arch:encoder}
\end{table}

\begin{table}[ht]
 \setlength{\tabcolsep}{0.5cm}
 \centering
 \begin{tabular}{c}
 \hline
 \hline
 $\z \in \R^{1 \times 1 \times nz}$  \\
 \hline
 Linear(nz, number of class) \\
 \hline
 \hline
 \end{tabular}
 \caption{Network architecture of the linear classifier.}
 \label{arch:linear_classifier}
\end{table}

\begin{table}[ht]
 \setlength{\tabcolsep}{0.5cm}
 \centering
 \begin{tabular}{c}
 \hline
 \hline
 $\z \in \R^{1 \times 1 \times nz}$  \\
  \hline
 Linear(nz, nz) ReLU \\
 \hline
 Linear(nz, number of clusters) \\
 \hline
 \hline
 \end{tabular}
 \caption{Network architecture of the MLP head for unsupervised conditional image generation}
 \label{arch:mlp_ucig}
\end{table}

\subsection{Optimization}
For all experiments, we use Adam \citep{kingma2014adam} as our optimizer, with hyperparameters $\beta_1 = 0.5, \beta_2=0.999$. The learning rate is set to be 0.0001. We choose $\epsilon^2=0.2$. For all experiments, we adopt augmentation from SimCLR \citep{chen2020simple}. 

For CIFAR-10, CIFAR-100, and Tiny ImageNet, we train our framework with a batch size of 1024 over 20,000 iterations. All experiments are conducted with at most 4 RTX 3090 GPUs. Methods that are compared against in Table \ref{tab:comparison_linearProbe_discriminative} are trained with the batch size of 256, because \citet{chen2020simple} observe that purely discriminative methods tend to perform better with smaller batch sizes.  Table \ref{tab:comparison_linearProbe_generative} methods have used their optimal parameters in their github code.

For training of the MLP head for unsupervised conditional image generation\eqref{eqn:cluster_mcr}, we again use Adam \citep{kingma2014adam} as our optimizer with hyperparameters $\beta_1 = 0.5, \beta_2=0.999$. We choose the learning rate to be 0.0001 and $\epsilon^2$ as 0.2, with batch size 1024 over 5000 iterations. 

For training of the linear classifier, we use Adam \citep{kingma2014adam} as our optimizer with hyperparameters $\beta_1 = 0.5, \beta_2=0.999$.  We choose learning rate to be 0.0001, with batch size 1024 over 5000 iterations.

\section{Additional Unsupervised Clustering and Generation Results}\label{sec:appendix_ucig}
\subsection{Cluster Reconstruction}
\begin{figure}[ht]
     \footnotesize
     \centering
     \subfigure[Cluster 1]{
         \includegraphics[width=0.485\textwidth]{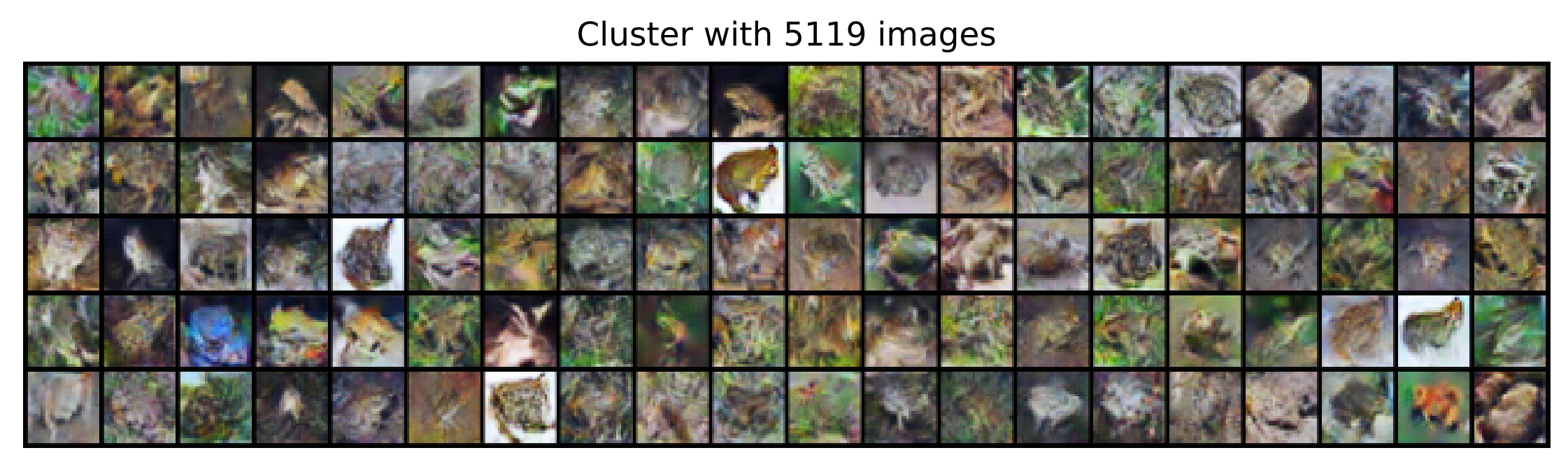}
     }
      \subfigure[Cluster 2]{
         \includegraphics[width=0.485\textwidth]{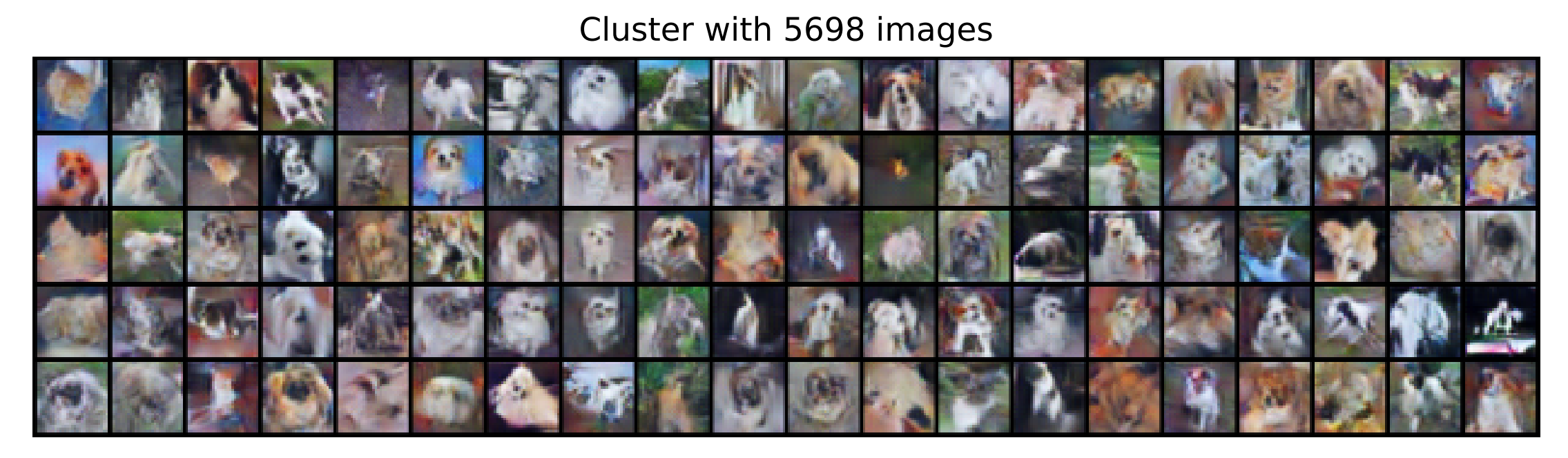}
     }
     \subfigure[Cluster 3]{
         \includegraphics[width=0.485\textwidth]{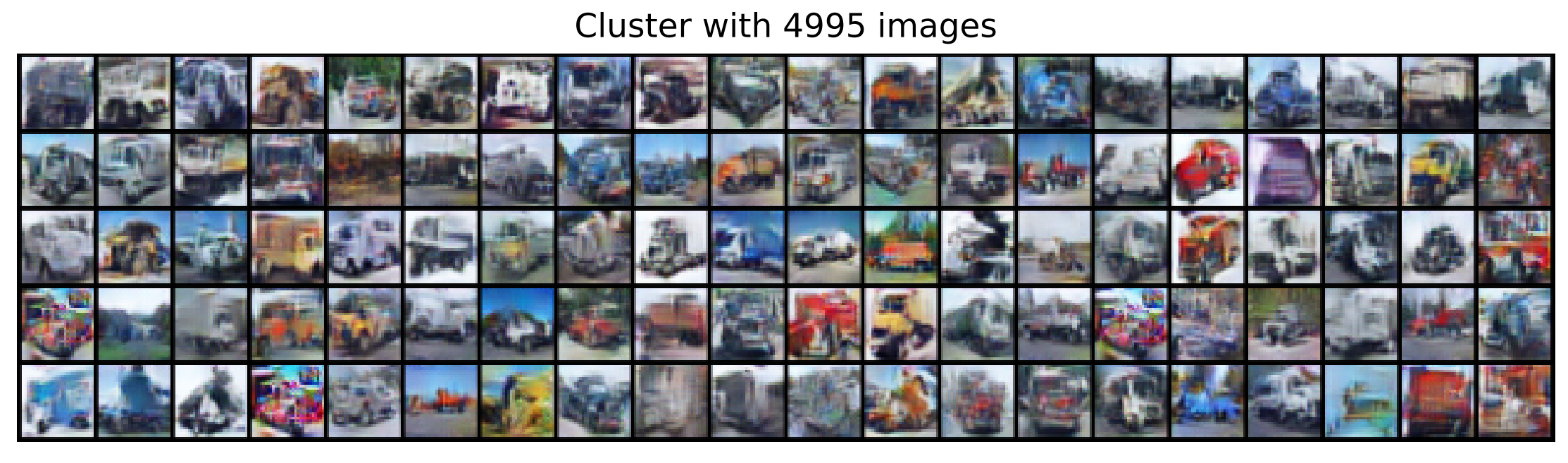}
     }
      \subfigure[Cluster 4]{
         \includegraphics[width=0.485\textwidth]{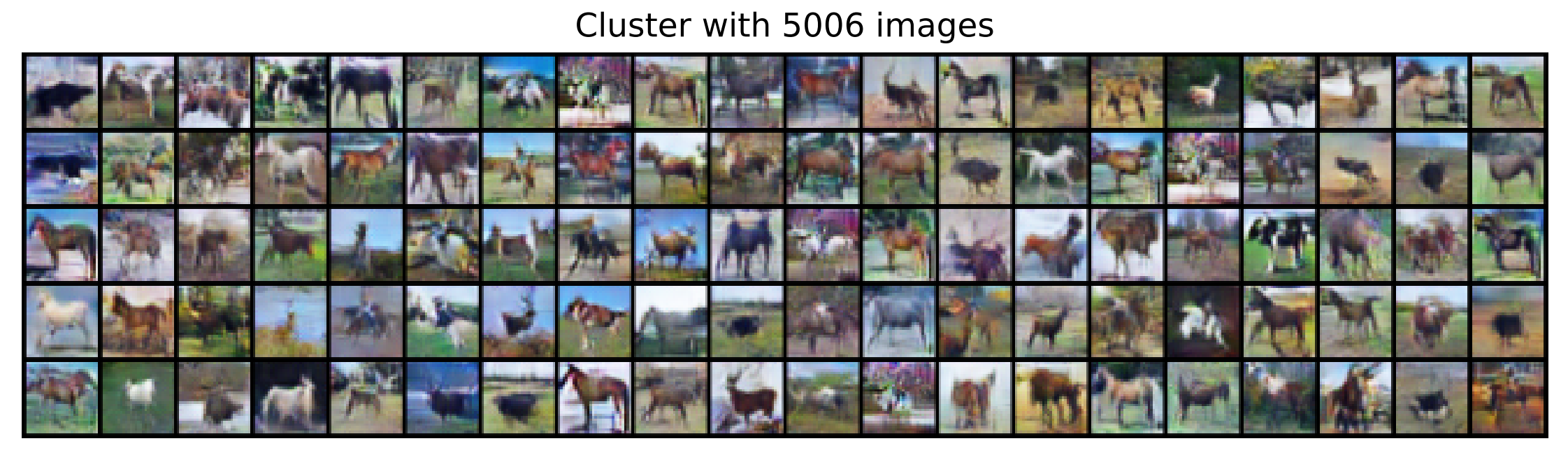}
     }
     \subfigure[Cluster 5]{
         \includegraphics[width=0.485\textwidth]{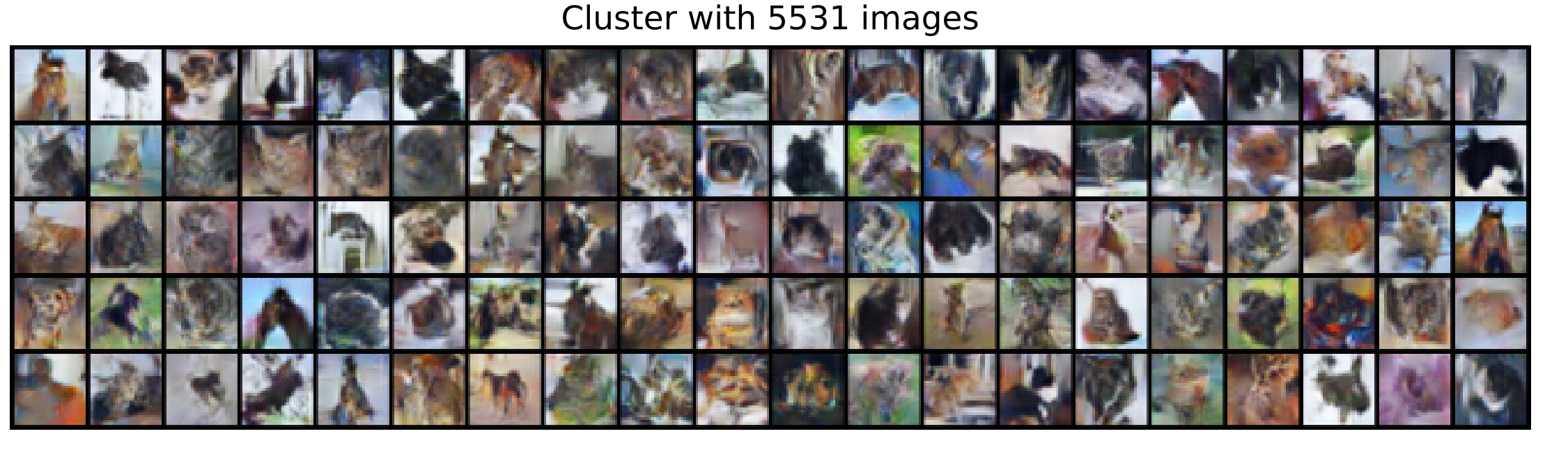}
     }
      \subfigure[Cluster 6]{
         \includegraphics[width=0.485\textwidth]{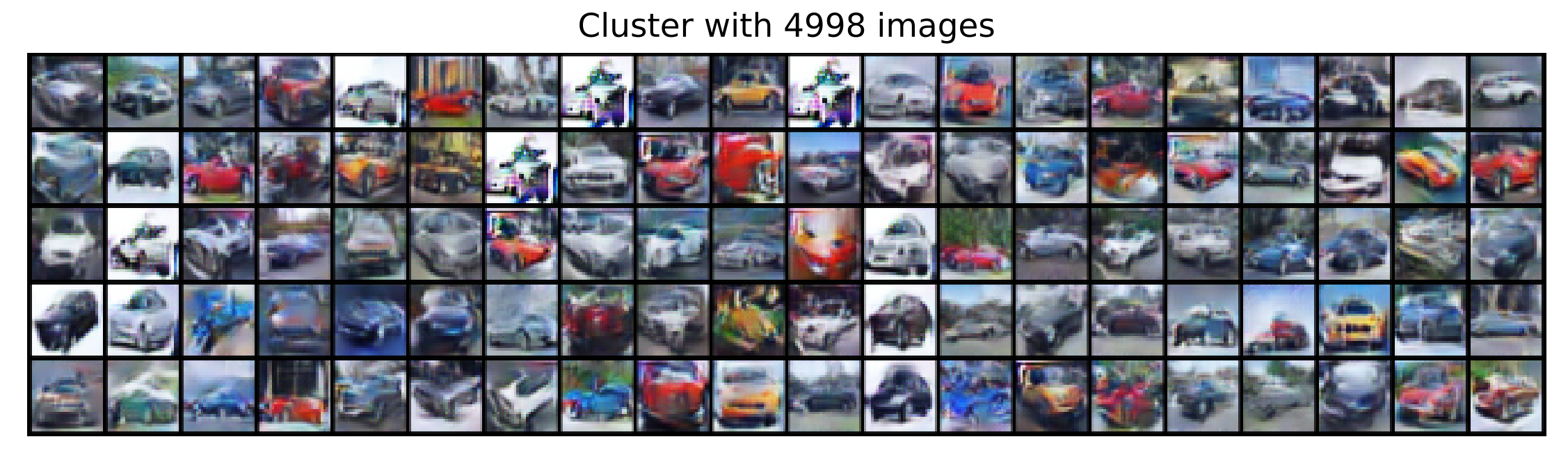}
     }
     \subfigure[Cluster 7]{
         \includegraphics[width=0.485\textwidth]{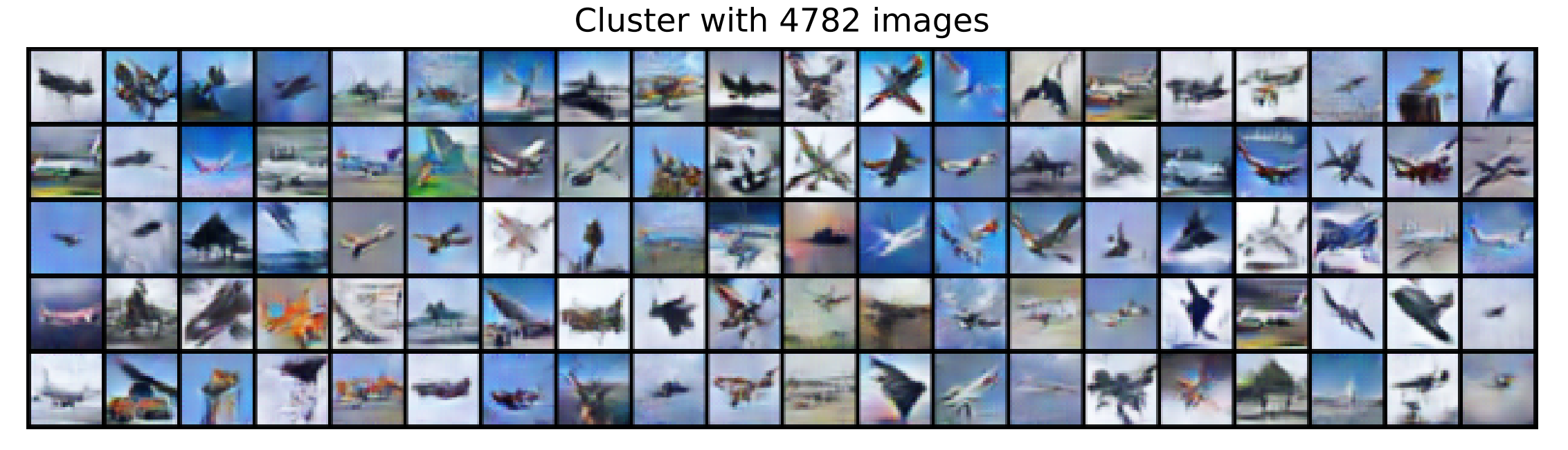}
     }
      \subfigure[Cluster 8]{
         \includegraphics[width=0.485\textwidth]{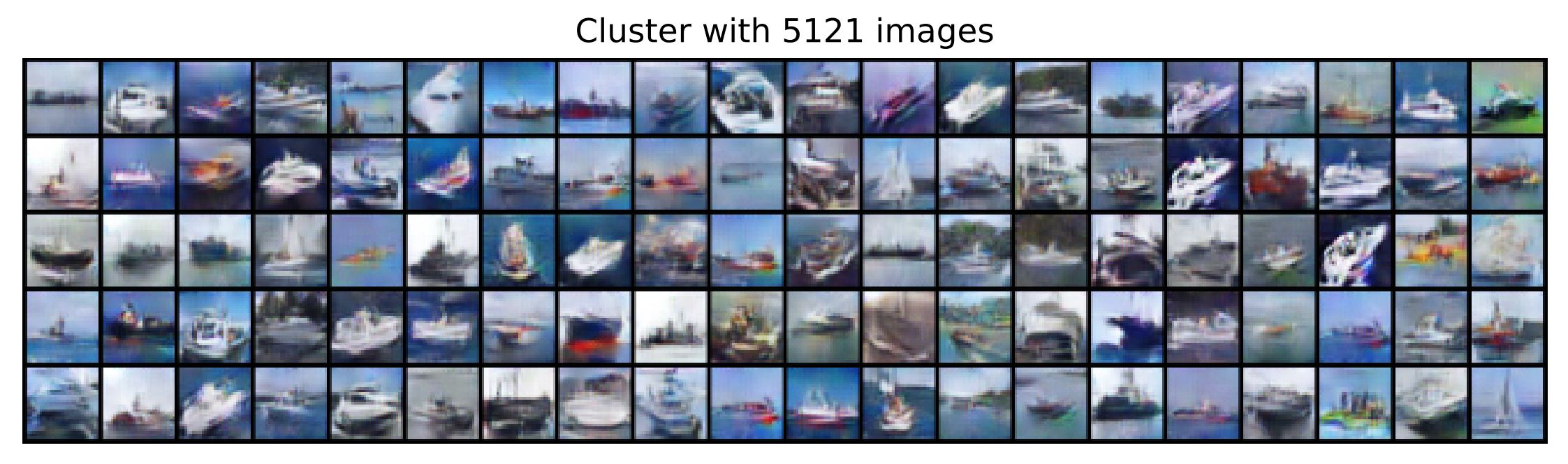}
     }
     \subfigure[Cluster 9]{
         \includegraphics[width=0.485\textwidth]{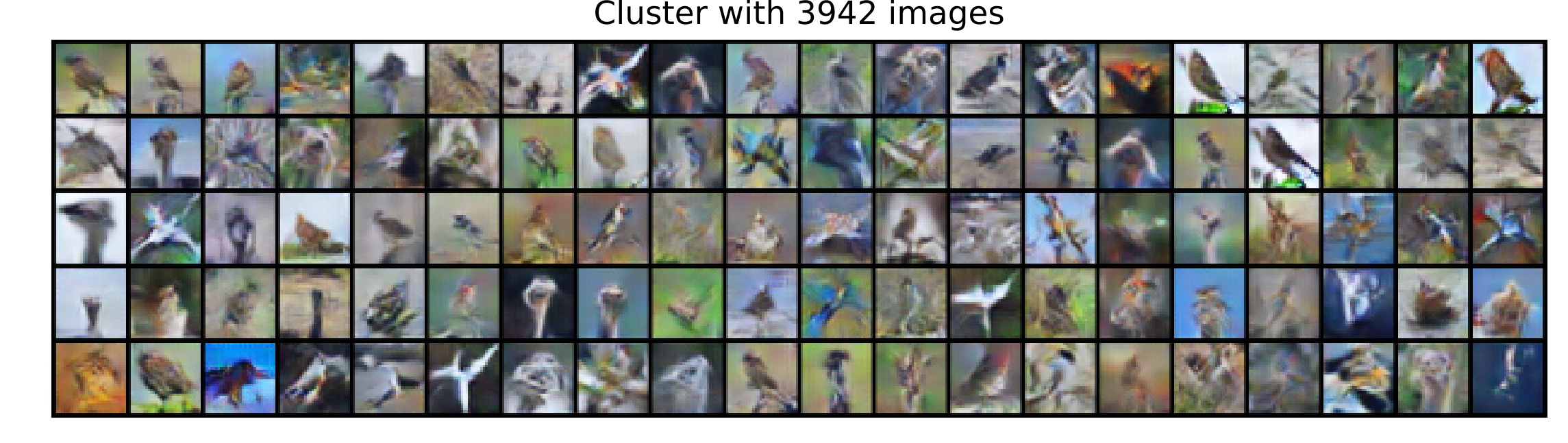}
     }
      \subfigure[Cluster 10]{
         \includegraphics[width=0.486\textwidth]{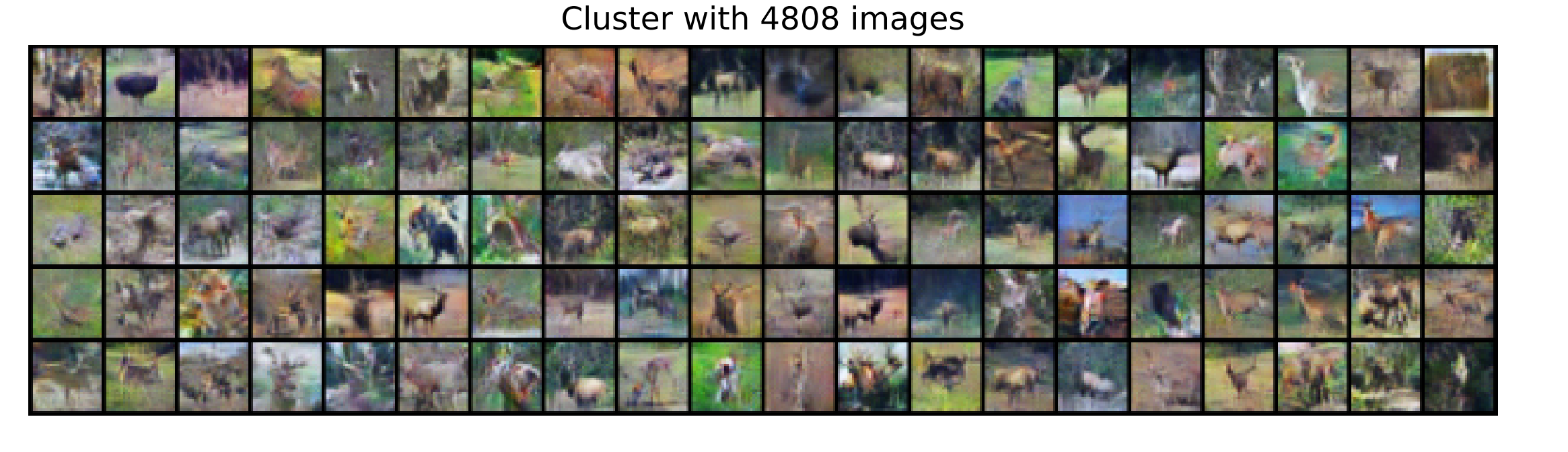}
     }
    \vspace{-0.15in}
    \caption{\small More result on the reconstruction of clusters in CIFAR-10 \vspace{-3mm}}
    \label{fig:recon_cluster}
\end{figure}

In this subsection, we visualize the reconstruction of ten clusters that are predicted and generated by \ours{} on the CIFAR-10 training set. Each block in Figure \ref{fig:recon_cluster} contains both a random sample of reconstructed data in a cluster and the total number of samples within it. Note that CIFAR-10 contains 50,000 training samples, split across 10 classes. As we see in Figure \ref{fig:recon_cluster}, the number of samples in each cluster are very close to 5,000, with the largest deviator (cluster 9) containing 3,942 samples. Without any cues, one can easily identify correspond each unsupervised cluster with a CIFAR-10 class. %tell the belonging of each cluster to the original CIFAR-10 classes. 
For a class like `bird', we observe that the model is able to group images of standing birds, flying birds, and bird heads, despite their visual differences.

\subsection{Unsupervised Conditional Image Generation}
\begin{figure}[ht]
     \footnotesize
     \centering
     \subfigure[Cluster 1]{
         \includegraphics[width=0.300\textwidth]{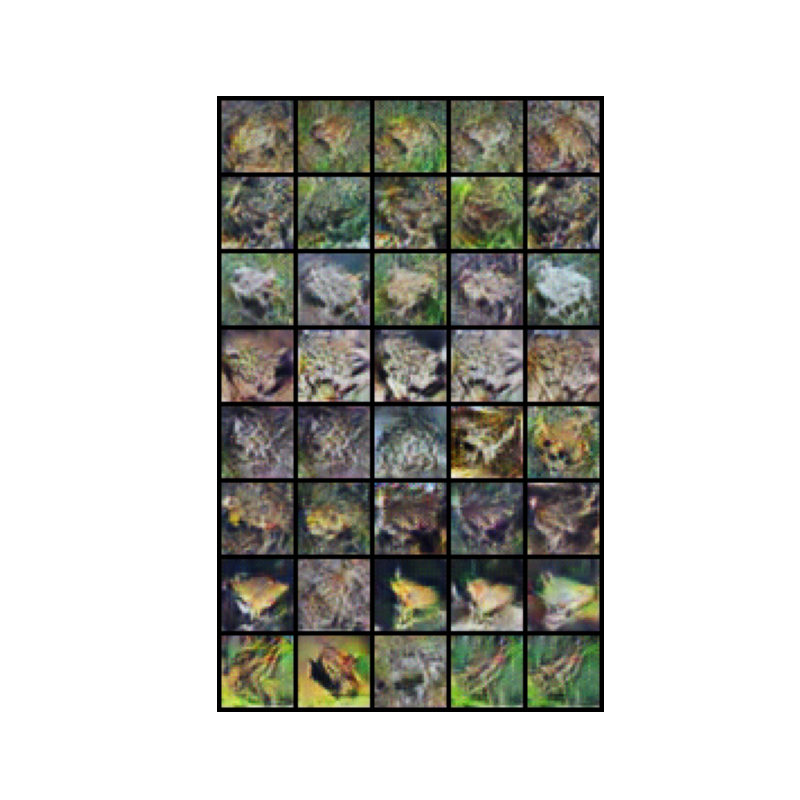}
     }
      \subfigure[Cluster 2]{
         \includegraphics[width=0.300\textwidth]{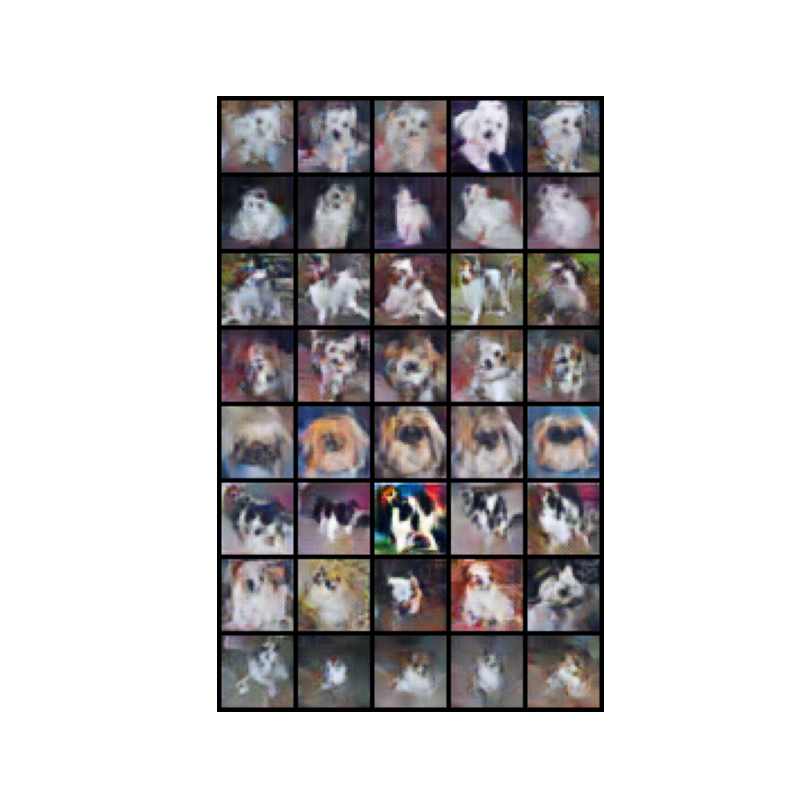}
     }
     \subfigure[Cluster 3]{
         \includegraphics[width=0.300\textwidth]{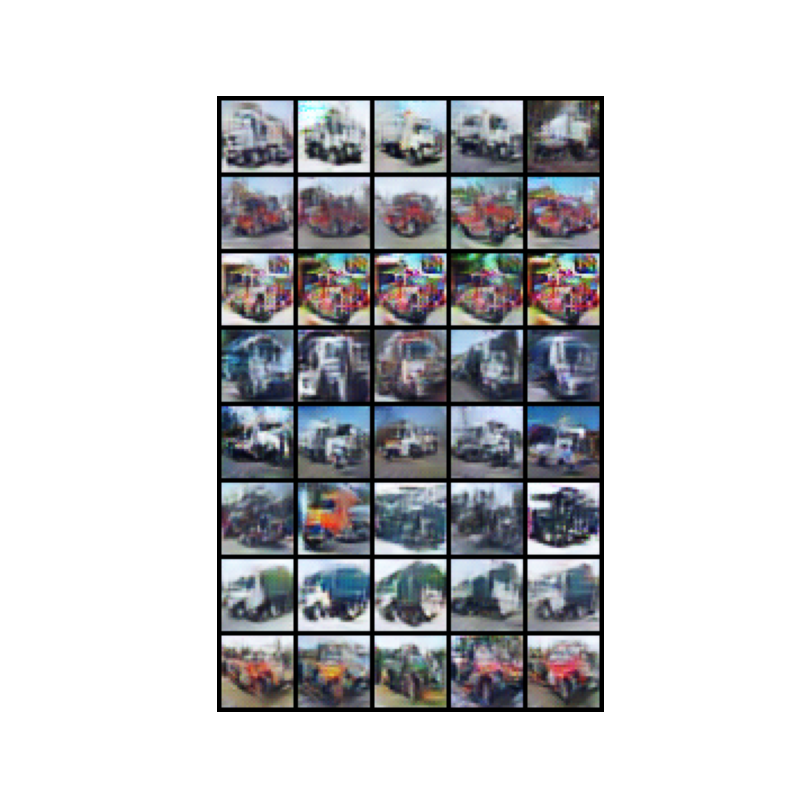}
     }
    \subfigure[Cluster 4]{
         \includegraphics[width=0.300\textwidth]{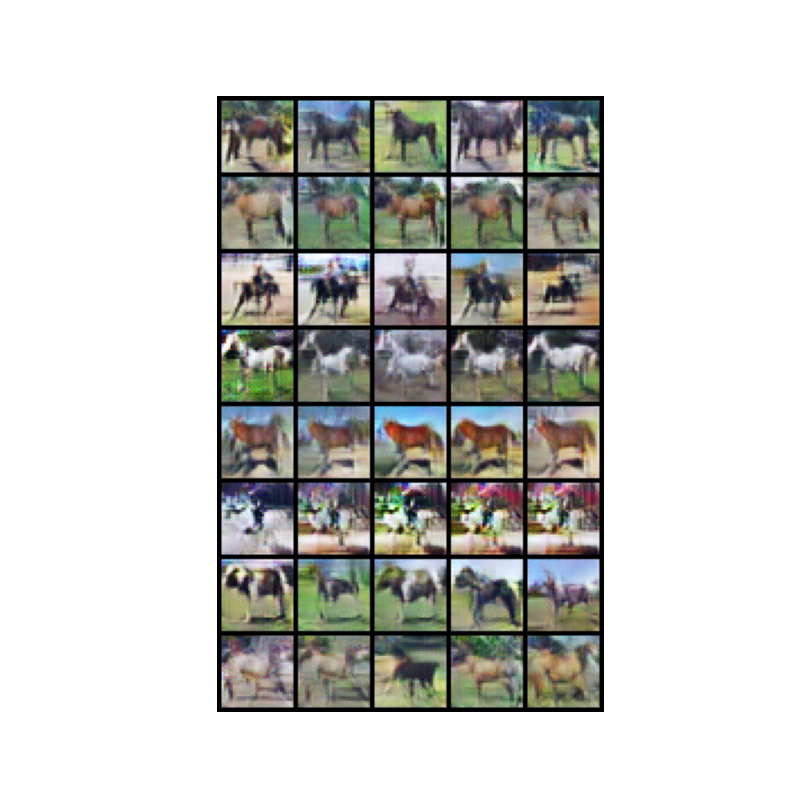}
     }
      \subfigure[Cluster 5]{
         \includegraphics[width=0.300\textwidth]{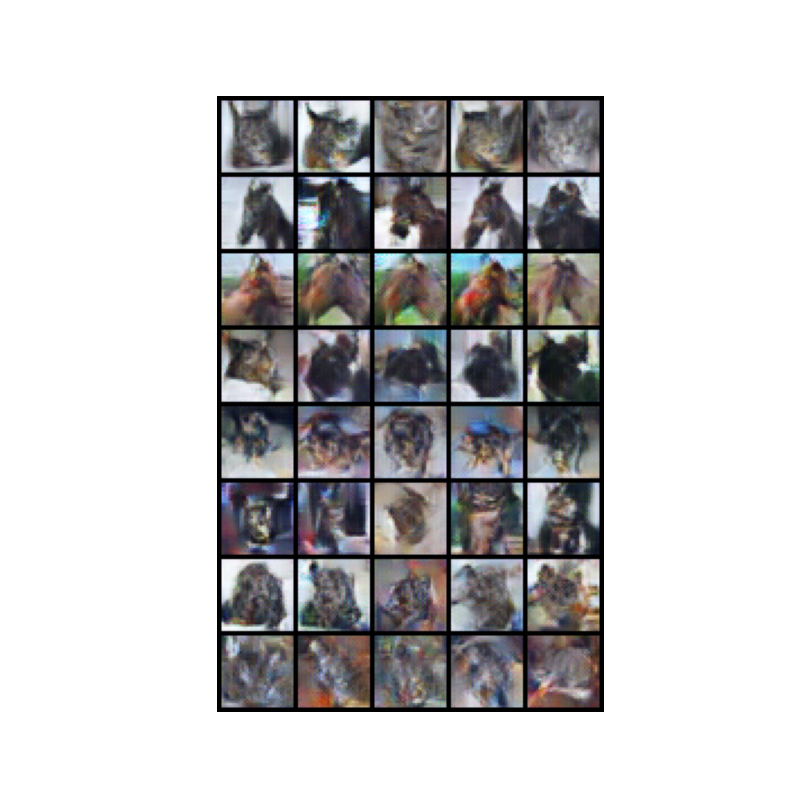}
     }
     \subfigure[Cluster 6]{
         \includegraphics[width=0.300\textwidth]{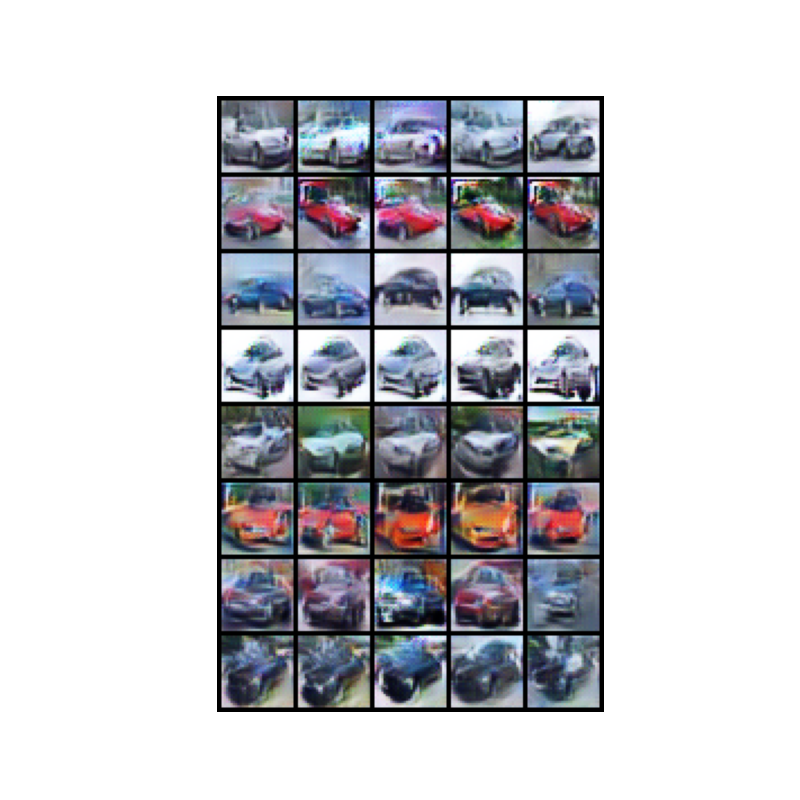}
     }
     \subfigure[Cluster 7]{
         \includegraphics[width=0.300\textwidth]{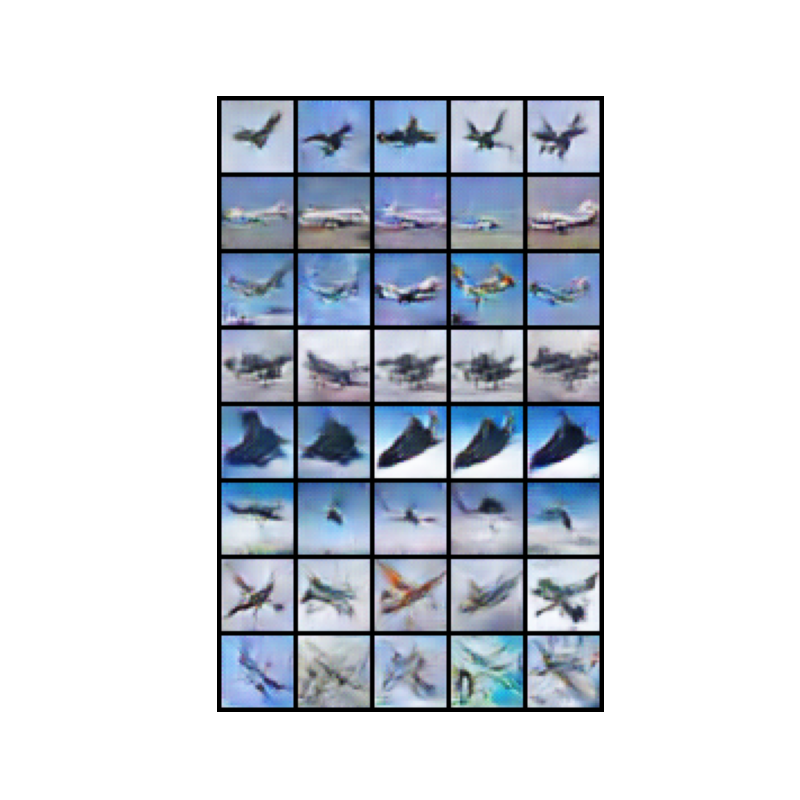}
     }
      \subfigure[Cluster 8]{
         \includegraphics[width=0.300\textwidth]{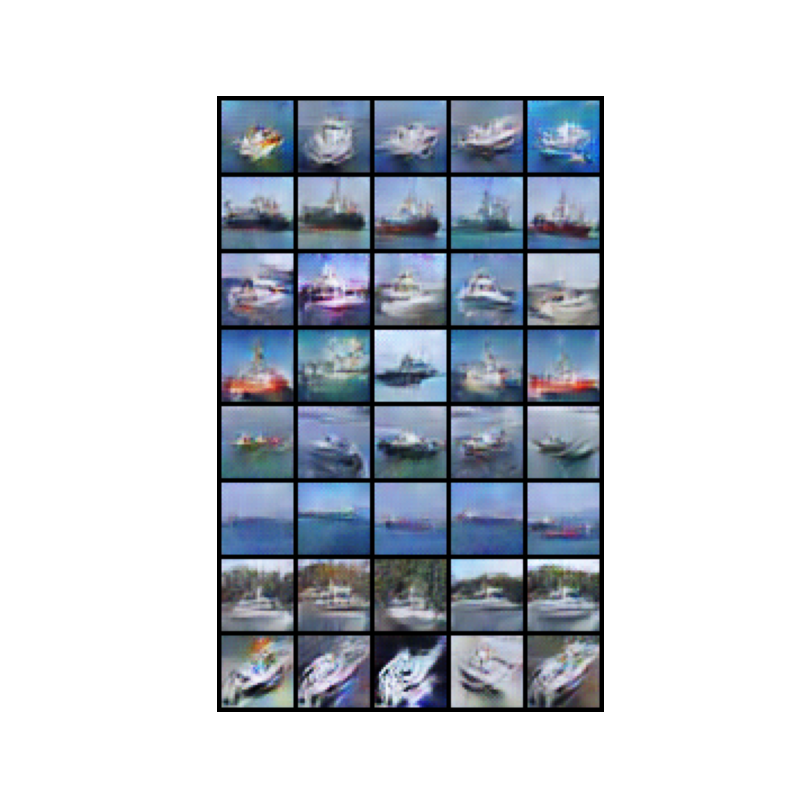}
     }
     \subfigure[Cluster 9]{
         \includegraphics[width=0.300\textwidth]{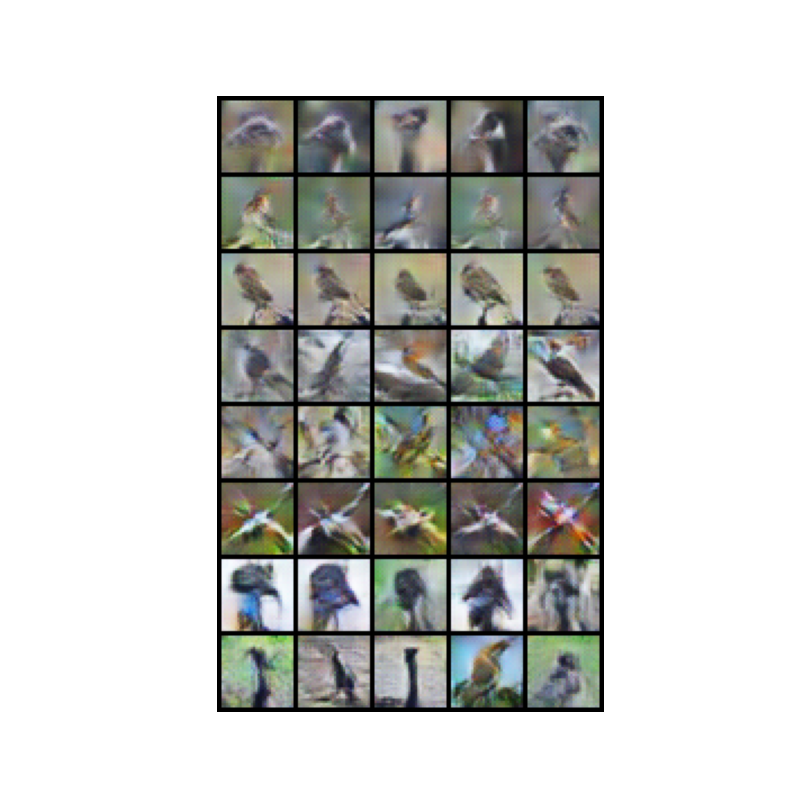}
     }
     \subfigure[Cluster 10]{
         \includegraphics[width=0.300\textwidth]{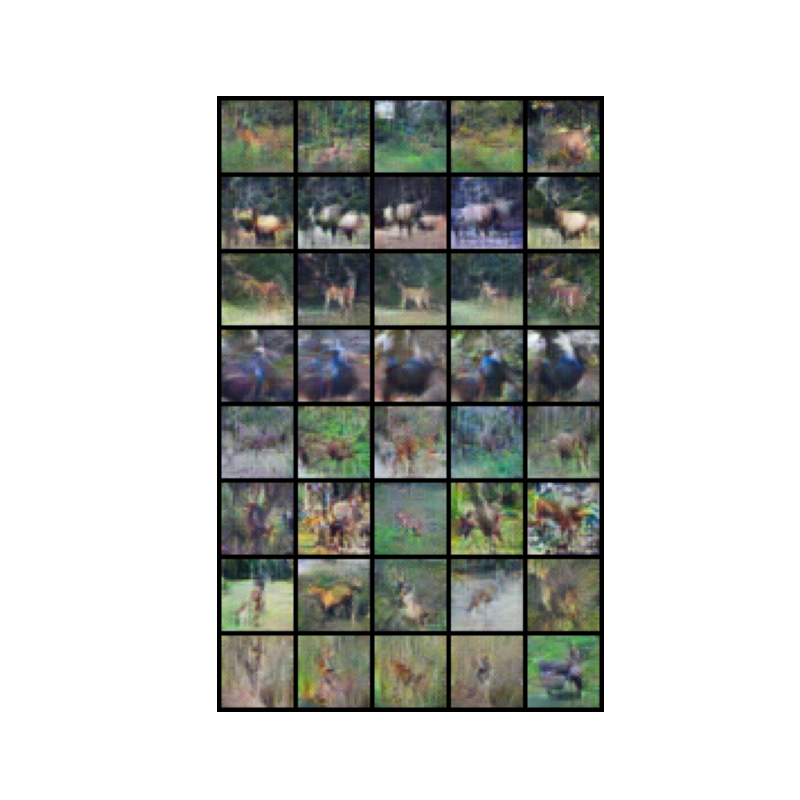}
     }
    \vspace{-0.15in}
    \caption{\small Unsupervised conditional image generation on CIFAR-10.
    Each block represents a cluster, within which each row represents one principal component direction in the cluster, and samples along each row represent different noises applied in that principal direction.
    \vspace{-3mm}}
    \label{fig:more_ucig}
\end{figure}

Building on \ours{}'s ability to cluster CIFAR-10 samples, we demonstrate the model's ability to perform unsupervised conditional image generation in Figure~\ref{fig:more_ucig}.
In contrast to reconstruction, where images are regenerated from features corresponding to real samples, we generate images based on the feature sampling technique proposed in \citep{dai2022ctrl}.
From these results, we observe that the \ours{} framework maintains in-cluster diversity, and that the diversity can be recovered and visualized via simple principal component analysis.

\section{Ablation Studies}
\subsection{The importance of each term in \ours{} formulation}
In this section, we study the significance of the sample-wise constraints and extra rate distortion term in the formulation \ref{eqn:constrained_maxmin}. Table \ref{tab::ablation_obj} presents the following objectives that we study:
\begin{itemize}[leftmargin=5mm]
    \item Objective I is the constrained \ours{} maximin \ref{eqn:constrained_maxmin}.
    \item Objective II is the constrained maximin without the augmentation compression constraint \ref{eqn:sample-compression}.
    \item Objective III is the constrained maximin without the sample-wise self-consistency constraint \ref{eqn:sample-self-consistency}.
    \item Objective IV is the constrained maximin without the extra rate distortion term.
    \item Objective V is the \ours{} without the augmentation compression constraint and sample-wise self-consistency constraint.
    \item Objective VI is the CTRL-Binary maximin formulation \ref{eqn:CTRL-Binary}.
\end{itemize}
Table \ref{tab:comparison_ablation} shows the result of a linear probe for representations trained using each objective on CIFAR-10. From the table, it is evident that both constraints and the rate distortion term are pivotal to the success of our framework.

\begin{table}[ht]
    \centering
    \small
    \begin{tabular}{l|l}
    ~ & ~ \\
    \hline
    Objective I:         &   $\max_\theta \min_\eta R(\Z) + \Delta R(\Z, \hat{\Z}) \mbox{ s.t.} \sum_{i\in N} \Delta R(\z^i, \hat{\z}^i) = 0, \mbox{and} \sum_{i\in N} \Delta R(\z^i, \z_{a}^i) = 0$ \\ 
    Objective II:         &   $\max_\theta \min_\eta R(\Z) + \Delta R(\Z, \hat{\Z}) \mbox{ s.t.} \sum_{i\in N} \Delta R(\z^i, \hat{\z}^i) = 0$ \\ 
    Objective III:         &   $\max_\theta \min_\eta R(\Z) + \Delta R(\Z, \hat{\Z}) \mbox{ s.t.} \sum_{i\in N} \Delta R(\z^i, \z_{a}^i) = 0$      \\
    Objective IV:         &   $\max_\theta \min_\eta \Delta R(\Z, \hat{\Z}) \mbox{ s.t.} \sum_{i\in N} \Delta R(\z^i, \hat{\z}^i) = 0, \mbox{and} \sum_{i\in N} \Delta R(\z^i, \z_{a}^i) = 0$      \\
    Objective V:         &   $\max_\theta \min_\eta R(\Z) + \Delta R(\Z, \hat{\Z})$      \\
    {Objective VI}:         &    {$\max_\theta \min_\eta \Delta R(\Z, \hat{\Z})$}      \\
    \end{tabular}
    \caption{Five different objective functions for  \ours{}.}
    \label{tab::ablation_obj}
\end{table}

\begin{table}[bht!]
\begin{small}
    \centering
    \setlength{\tabcolsep}{4.5pt}
    \renewcommand{\arraystretch}{1.25}
    \begin{tabular}{l|c|c|c|c|c|c}
    Method  & Objective I & Objective II  & Objective III & Objective IV& Objective V& Objective VI\\
    \hline
    \hline
    Accuracy    & 0.874 & 0.578  & 0.644 & 0.522 & 0.633 & 0.599  \\  
    \end{tabular}      
    % \vspace{-0.1in}
    \caption{\small Ablation study on the significance of different terms in \ours{}.\vspace{-7mm}}
     \label{tab:comparison_ablation}
\end{small}     
\end{table}

% $ R(\Z) + \Delta R(\Z, \hat{\Z}) \label{eqn:constrained_maxmin}\\
%  \mbox{subject to} \quad & \sum_{i\in N} \Delta R(\z_{conv}^i, \hat{\z}_{conv}^i) = 0, \;\; \mbox{and} \;\; \sum_{i\in N} \Delta R(\z^i, \z_{a}^i) = 0$

\subsection{The importance of MCR$^2$ in \ours{} formulation}

\begin{figure}[bht!]
     \footnotesize
     \centering
     \subfigure[CIFAR-10 $\X$]{
         \includegraphics[width=0.315\textwidth]{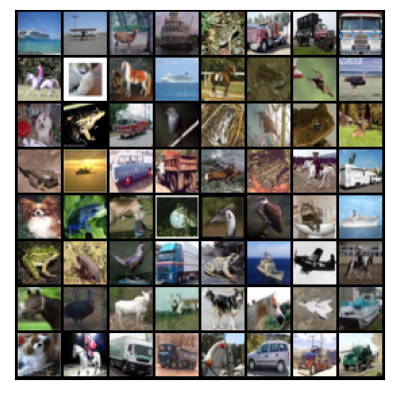}
     }
     \subfigure[CIFAR-10 $\hat{\X}$]{
         \includegraphics[width=0.315\textwidth]{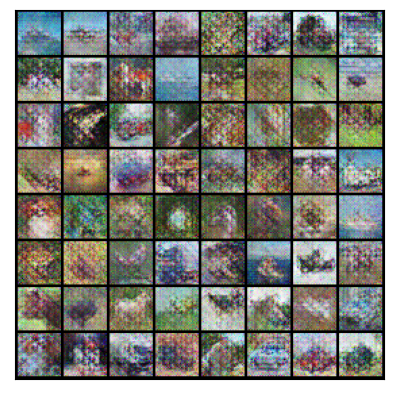}
     }
    \vspace{-0.15in}
    \caption{\small Visualization of images trained by U-CTRL-noMCR$^2$: $\X$ and reconstructed $\hat{\X}$ on CIFAR-10 dataset. \vspace{-6mm}}
    \label{fig:vis_recon_ablation}
\end{figure}

In this section, we verify the significance of MCR$^2$ term $\Delta R(\Z,\hat{\Z})$ in our method. We do ablation study on CIFAR-10 with the same network and training condition. If we take away MCR$^2$ from our formulation, it changes \eqref{eqn:ablation_constrained_maxmin}. For simplicity, we call it U-CTRL-noMCR$^2$
\begin{align}
      \max_\theta \min_\eta  \quad & R(\Z) \label{eqn:ablation_constrained_maxmin}\\
 \mbox{subject to} \quad & \sum_{i\in N} \Delta R(\z^i, \hat{\z}^i) = 0, \;\; \mbox{and} \;\; \sum_{i\in N} \Delta R(\z^i, \z_{a}^i) = 0. \nonumber
\vspace{-2mm}
\end{align}
Table \ref{tab:comparison_ablation_mcr2} shows that \ours{} without the MCR$^2$ not only learns worse representation but also generalizes worse to out of distribution data. Figure \ref{fig:vis_recon_ablation} visualizes the reconstructed $\hat{\X}$ by U-CTRL-noMCR$^2$. It is clear from the image figure that without the MCR$^2$, the decoder fails to reconstruct high-quality images. 
\begin{table}[bht!]
    \centering
    \setlength{\tabcolsep}{4.5pt}
    \renewcommand{\arraystretch}{1.25}
    \begin{tabular}{l|l|ll}
                             & Accuracy on CIFAR-10 & Transfer Accuracy on CIFAR-100 &   \\ 
    \cline{1-3}
    \ours{} & 0.874                & 0.481                          &   \\ 

    U-CTRL-noMCR$^2$    & 0.836                & 0.418                          &  
    \end{tabular}
    \caption{\small Ablation study on the significance of MCR$^2$ in \ours{}.\vspace{-7mm}}
     \label{tab:comparison_ablation_mcr2}
\end{table}

It follows our discussion in the introduction that discriminative tasks and generative tasks together learn feature that benifits each other.

\section{Random Seed Sensitivity}
In this section, we verify the stability of our method against different random seeds. 
We report in Table \ref{tab:ablation_random_seed} the accuracy of \ours{} on CIFAR-10 with different seeds. We observe that the choice of seed has very little impact on performance. 
\begin{table}[bht!]
\begin{small}
    \centering
    \setlength{\tabcolsep}{5pt}
    \renewcommand{\arraystretch}{1.25}
    \begin{tabular}{l|c|c|c|c|c}
    % \hline
    Random Seed   & 1 & 5 & 10 & 15 & 100 \\
    \hline
    \hline
    Accuracy    & 0.874 & 0.876 & 0.870 & 0.874  & 0.871        \\  
    % \hline
    \end{tabular}      
    %\vspace{-0.2in}
    \caption{Ablation study on varying random seeds.}
     \label{tab:ablation_random_seed}
     \end{small}
\end{table}

\end{document}